\documentclass{article} 
\usepackage{iclr2026_conference,times}

\usepackage{amsmath,amsfonts,bm}

\def\eqref#1{equation~\ref{#1}}

\def\1{\bm{1}}

\DeclareMathAlphabet{\mathsfit}{\encodingdefault}{\sfdefault}{m}{sl}
\SetMathAlphabet{\mathsfit}{bold}{\encodingdefault}{\sfdefault}{bx}{n}

\usepackage{hyperref}
\usepackage{url}
\usepackage{graphicx}
\usepackage{float}
\usepackage{wrapfig} 

\usepackage{booktabs}
\usepackage{makecell}
\usepackage[table]{xcolor}
\usepackage{scalerel}
\newcommand{\smallcheck}{{\scaleto{\checkmark}{4pt}}}
\newcommand{\method}{LT}
\newcommand{\methodfull}{Latent-Trajectory}
\usepackage{enumitem}
\usepackage{multirow}
\usepackage{nicematrix}
\usepackage{siunitx}

\usepackage{authblk}

\title{\emph{Tracing the Traces}: Latent Temporal Signals for Efficient and Accurate Reasoning}

\iclrfinalcopy 

\makeatletter
\g@addto@macro\maketitle{%
  \thispagestyle{fancy}
  \fancyhead[L]{}
}
\makeatother

\begin{document}


\author{%
\textbf{Martina G. Vilas}\textsuperscript{1,2}\thanks{Work done during internship at Microsoft Research. Correspondence to martinagvilas@em.uni-frankfurt.de and vidhishab@microsoft.com}\quad
\textbf{Safoora Yousefi}\textsuperscript{2}\quad
\textbf{Besmira Nushi}\textsuperscript{3}\thanks{Work done at Microsoft Research}\quad
\textbf{Eric Horvitz}\textsuperscript{2}\par
\textbf{Vidhisha Balachandran}\textsuperscript{2}
\par
\vspace{0.6em}
\textsuperscript{1}Goethe University Frankfurt\quad
\textsuperscript{2}Microsoft Research\quad
\textsuperscript{3}NVIDIA
}

\maketitle
\begin{abstract}
Reasoning models improve their problem-solving ability through inference-time scaling, allocating more compute via longer token budgets. 
Identifying which reasoning traces are likely to succeed remains a key opportunity: reliably predicting productive paths can substantially reduce wasted computation and improve overall efficiency. 
We introduce \emph{Latent-Trajectory} signals that characterize the temporal evolution of a model's internal representations during the generation of intermediate reasoning tokens. 
By measuring the overall change in latent representations between the start and end of reasoning, the change accumulated across intermediate steps, and the extent to which these changes advance toward the final state, we show that these signals predict solution accuracy more reliably than both cross-layer metrics and output-based confidence measures.
When used to guide answer selection across multiple sampled generations, Latent-Trajectory signals make test-time scaling more effective and efficient than majority voting, reducing token usage by up to $70$\% while preserving and even improving accuracy by $2.6$\% on average. 
Moreover, these predictive signals often emerge early in the reasoning trace, enabling early selection and allocation of compute to the most promising candidates.
Our findings contribute not only practical strategies for inference-time efficiency, but also a deeper interpretability perspective on how reasoning processes are represented and differentiated in latent space.
\end{abstract}

\section{Introduction}

Recent advances in large language models (LLMs) have shown that complex reasoning tasks can be solved more effectively by scaling compute at inference time to generate longer and multiple chains-of-thought (\textit{reasoning traces}) and aggregating them into a final solution \citep{guo2025deepseek, abdin2025phi, openai2024reasoning, yang2025qwen3}.
However, not all reasoning traces are equal: while some contain productive steps that lead to correct answers, others may deviate into unproductive paths such as overthinking, failing to converge on a valid solution strategy, or exhibiting inconsistent reasoning  \citep{shojaee2025illusion, chen2024overthink, sun2025omega}. 
Identifying the quality of a reasoning trace (the likelihood of it leading to a correct solution) is critical. 
It not only enables more reliable prediction of correct answers, but it can also improve computational efficiency by potentially avoiding wasted effort on unproductive paths, and can provide feedback signals that can enhance model training. 
By understanding which reasoning processes are effective, we can systematically guide models toward reinforcing productive strategies and suppressing ineffective ones. 

Prior work has approached this problem by inspecting reasoning traces in their surface natural-language form and identifying behaviors that lead to accurate answers, an approach that typically relies on costly human or model annotations \citep{lee2025cot, gandhi2025cognitive}. 
In addition, natural language reasoning traces may not always reflect the underlying strategies that models employ \citep{chen2025faithfulness, stechly2025beyond}, and some models are trained to produce intermediate latent embeddings rather than explicit text \citep{hao2024continuous}.
Thus, language alone may be an unreliable proxy for evaluating reasoning trace quality.
Other work has explored heuristic signals like trace length \citep{hassid2025shorter, marjanovic2025thoughtology}, output distribution statistics \citep{kadavath2022language, yona2022useful}, agreement-based self-consistency \citep{wang2023selfconsistency}, or using trained verifiers \citep{li-etal-2023-making, zhang-etal-2024-small} to identify correct solutions, but these methods often trade accuracy for simplicity, or computational cost for accuracy.

We explore an alternative direction that solely leverages a models' trajectory of hidden states to predict which traces lead to a correct solution. 
Previous studies have shown that probing hidden states can reveal informative signals about safety \citep{turner2023steering, zou2023representation}, learning dynamics \citep{olsson2022incontext, hosseini2023straight}, reliability \citep{meng2022locating, Yuksekgonul2024AttentionSatisfies}, and performance \citep{wang2024latent} of LLMs.
Building on this perspective, we hypothesize that the temporal evolution of hidden states during the generation of intermediate reasoning tokens contains predictive information about the final solution correctness, and can be leveraged for more compute-efficient and accurate inference.

\begin{wrapfigure}{r}{0.55\textwidth}
\vspace{-1.5em} 
\centering
\includegraphics[width=0.53\textwidth,trim=10 0 10 10,clip]{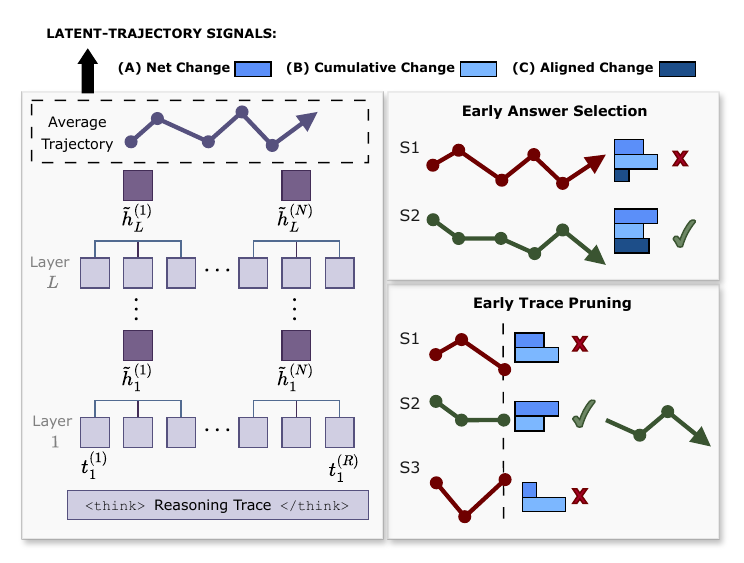}

\vspace{-1.5em}
\caption{Latent-Trajectory framework. Trajectory vectors are constructed from token-level hidden states, and a set of three signals is derived to quantify their temporal evolution. These signals predict successful traces and enable answer selection and early path selection in multi-sample inference.}
\vspace{-0.9em}
\label{fig:fig1}
\end{wrapfigure}
We introduce a family of \emph{Latent-Trajectory} (LT) signals that capture three complementary temporal aspects of a model's internal representational trajectory (see Figure \ref{fig:fig1}): (i) the total representational change from the start to the end of the trace, (ii) the change accumulated across intermediate steps, and (iii) the extent to which the intermediate updates progress towards or away from the final state.
These metrics operate directly on hidden states, require no additional training or external annotations, and can be computed during inference.

Our experiments evaluate the use of LT signals across families of reasoning-enabled LLMs (DeepSeek-R1-Distill-Qwen14B, Phi4-Reasoning-Plus, Qwen3-14B) and domains spanning science, math, and path optimization problems. 
We show that LT can reliably distinguish between traces leading to correct versus incorrect answers, yielding significantly higher discriminatory power than methods using other model internal or output-distribution-based signals. 
At inference-time, \method{} can be leveraged to achieve both higher efficiency and improved accuracy.
In sample scaling experiments, early answer selection using LT yields up to a $70$\% reduction in token usage, along with $2.6$\% average accuracy gain over majority-vote baselines by reducing the number of generations sampled. 
In addition, these signals often emerge early in the trace, enabling early recognition of strong candidates and allocating compute to them.

Overall, we show that a model’s internal dynamics can be reliable predictors of reasoning quality, offering both practical tools for inference-time control and interpretability insights into how reasoning trajectories evolve, opening paths for broader applications that exploit internal signals for efficiency, accuracy, and calibrated decision-making.

\vspace{-2ex}
\section{Related Work}
\textbf{Assessing Reasoning Quality:} 
A growing body of work seeks to quantify the quality of reasoning traces in order to predict solution accuracy with high reliability. 
Many strategies involve employing verifier models, external or self, to assess the correctness of candidate answers \citep{weng-etal-2023-large, NEURIPS2023_91edff07, zhang-etal-2024-small}. 
These approaches are effective but substantially increase inference cost. 
An alternative direction performs fine-grained analyses of the trace surface form, proposing metrics that target factual and logical validity, as well as linguistic and semantic coherence \citep{wu2025knowledge, golovneva2022roscoe}. 
Heuristics derived from output token distributions or from trace length have also been explored to decide whether a path is likely to be accurate \citep{hassid2025shorter, kadavath2022language, yona2022useful}. 
Such methods often require annotation or structured extraction from traces, which introduces dependence on human raters or auxiliary expert models and can lead to model-specific heuristics. 
In contrast, \method{} signals are computed directly at inference time without a teacher model or additional runs, which yields a more efficient procedure. 
Concurrent work trains model-specific probes over hidden representations to detect when intermediate answers are likely correct \citep{zhang2025reasoning}. Our approach shares the objective but remains training-free and can be applied to diverse models and datasets with minimal setup.

\textbf{Representational Analysis:} 
Previous studies have shown that probing an LLM's hidden states reveals informative signals about reliability \citep{meng2022locating, Yuksekgonul2024AttentionSatisfies}, safety \citep{turner2023steering, zou2023representation}, performance \citep{wang2024latent}, and learning dynamics \citep{olsson2022incontext, hosseini2023straight}. 
We extend this research direction to leverage hidden states to predict solution correctness in reasoning models.
Closest to our approach, \cite{wang2024latent} examines representational curvature across layers (i.e. spatial perspective) for predicting accuracy in instruction-tuned models. 
We differ by adopting a temporal perspective across tokens and focusing specifically on reasoning models.
Concurrent work \citep{li2025core} extends sequential representational analysis to detect repetition loops in mathematical reasoning, further supporting the premise that temporal latent dynamics provide valuable insight into model behavior.

\textbf{Efficient Inference Scaling:} 
Scaling up inference-time computation is a key factor in improving reasoning performance in LLMs \citep{openai2024reasoning, guo2025deepseek, abdin2025phi}. 
While effective, previous studies \citep{balachandran2025inference, shojaee2025illusion, sui2025stop_tmlr} show that models trained to generate long reasoning traces exhibit ‘overthinking’ and consume compute even after reaching a correct solution. 
This has motivated efforts to curb such behavior, either by training models to produce more concise reasoning \citep{kang2025c3ot, shrivastava2025sample} or by dynamically halting trace generation once the model is confident in its answer \citep{yang2025dynamic, zhang2025reasoning}. 
Another inference-time scaling strategy is to generate multiple samples and aggregate answers using self-consistency \citep{wang2023selfconsistency}, external verifiers \citep{zhang-etal-2024-small}, or iterative self-verification \citep{NEURIPS2023_91edff07, balachandran2025inference}. 
These methods boost accuracy for both standard and reasoning models but come with substantially higher computational cost.
Recent work has sought to improve efficiency by pruning reasoning paths with trained classifiers \citep{manvi2024adaptive, li2024escape}. 
In contrast, our experiments show that Latent-Trajectory signals provide a training-free way to guide reasoning-path selection and answer aggregation.

\section{Latent-Trajectory Signals of Reasoning Quality}
\label{method}

Given a problem, reasoning models generate a sequence of tokens composed of a reasoning trace followed by a final answer. 
The trace is often delimited by special tokens ($\{trace\_start\}$, $\{trace\_end\}$), such that:
\[
q_1, \dots, q_i \; \texttt{\{trace\_start\}} \; t_1, \dots, t_r \; \texttt{\{trace\_end\}} \; a_1, \dots, a_j,
\]
where \( q_1, \dots, q_i \) are the user query (problem) tokens, \( t_1, \dots, t_r \) are the reasoning trace tokens, and \( a_1, \dots, a_j \) are the final answer tokens.
For each position \( r \in \{1, \dots, R\} \) within the reasoning trace, the model produces a hidden state of activations at each layer \( l \in \{1, \dots, L\} \), denoted by \( h_l^{(r)} \in \mathbb{R}^d \). These hidden states form a 2D array of $d$-sized representations, indexed by layer and token position, and encode the \emph{latent space} of the model at each step of the reasoning trace (see Figure \ref{fig:fig1}).

\subsection{Latent-Trajectory signals}

We aim to assess the quality of a model’s intermediate reasoning by analyzing how its internal representations evolve throughout the reasoning trace.
To quantify this, we average token-level activations into a sequence of segment-level states and extract trajectory signals that quantify the magnitude and geometry of representational change.

First, to enhance the robustness of the signal and reduce dimensionality, we divide the reasoning trace \( t_1, \dots, t_r \) into non-overlapping \emph{reasoning segments}, where each segment is a contiguous block of $k$ tokens ($k$ = 500)\footnote{We experimented with various segmentation methods, including delimiters. See Appendix \ref{appendix_averaging}.}. 
For each transformer layer $l \in {1,\dots,L}$ and segment index $n \in {1,\dots,N}$, we compute the segment-level hidden state $\tilde{h}^{(n)}_l$ by averaging the token hidden states within that segment.
Intuitively, $\tilde{h}^{(n)}_l$ corresponds to the average representation the model maintains in latent space while processing segment $n$.
This temporal coarse-graining smooths local fluctuations in token-level dynamics while preserving the large-scale evolution of the model’s latent space over the trace. The sequence $\{\tilde{h}^{(1)}_l,\dots,\tilde{h}^{(N)}_l\}$ at layer $l$ provides a trajectory-level encoding of the hidden-state evolution over the intermediate reasoning tokens.

\begin{wrapfigure}{r}{0.40\textwidth}
\vspace{-1.8em} 
\centering
\includegraphics[width=0.38\textwidth,trim=20 0 20 0,clip]{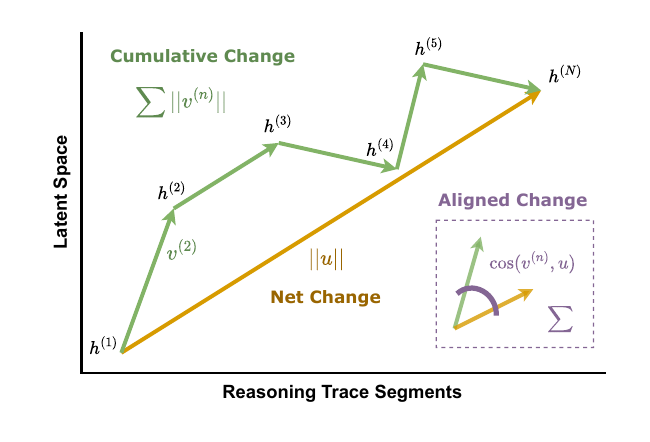}
\vspace{-2em} 
\caption{Latent-Trajectory signals.}
\label{metrics}
\vspace{-0.9em} 
\end{wrapfigure}
Given the segment hidden states, we define two basic vectors that anchor our signals. The \emph{reasoning drift vector}:
$
u_l = \tilde{h}^{(N)}_l - \tilde{h}^{(1)}_l,
$
captures the overall direction and distance the model’s internal state travels during the trace.
Complementarily, the \emph{update vector} for segment $n$:
$
v^{(n)}_l = \tilde{h}^{(n)}_l - \tilde{h}^{(n-1)}_l, \quad n=2,\dots,N
$
describes the incremental change between consecutive reasoning segments. 
Taken together, ${u_l}$ and ${v^{(n)}_l}$ capture not only the overall extent of representational movement, but also the step-by-step dynamics of how that movement unfolds.

From these primitives we derive three complementary signals that summarize (i) overall representational change, (ii) accumulated change over the trace, and (iii) extent of progress towards the final state (see Figure \ref{metrics}). 
Each signal is computed per layer and then averaged across layers to yield a single score per trace.

\paragraph{Net Change.}
First, we explore whether intermediate reasoning substantially alters the model’s latent space and whether such change is predictive of accuracy.
To assess this, we measure the magnitude of representational change in the latent space between the first and last reasoning segment.

Formally, we measure the norm of the drift vector $u_l$ at each layer, which encodes the magnitude of this change, and normalize by the number of segments to control for trace length.
Finally, we average across layers to obtain a single score:
\[
\textsc{NetChange} = \frac{1}{L} \sum_{l \in L} \frac{\|u_l\|_2}{N}
\]
Larger values indicate that the final hidden state has substantially changed from the initial state, suggesting that the reasoning steps produced significant changes in representational space.

\paragraph{Cumulative Change.}
While Net Change measures the overall representational change between the initial and final reasoning segments, it does not characterize the intermediate latent-space changes.
To summarize the total amount of representational movement along the trace, we additionally compute the cumulative magnitude of the sequential updates to the reasoning trace.

We consider the update vectors $v^{(n)}_l$, which represent the changes in layer $l$ between consecutive reasoning segments. 
The norm of the update vectors $||v^{(n)}_l||_2$ gives the magnitude of change at each step, and aggregating the norms across all segments captures the total movement along the trajectory. 
Finally, averaging across layers yields a single score:
\[
\textsc{CumulativeChange} \;=\; \frac{1}{L}\sum_{l \in L} \sum_{n=2}^{N} \|v^{(n)}_l\|_2
\]

Intuitively, Cumulative Change quantifies the overall shifts in representations during the course of reasoning, independent of the final states. 
Larger values encode significant variations in representations across segments, while smaller values encode negligible or incremental updates.

\paragraph{Aligned Change.} 
Beyond measuring the magnitudes of overall and intermediate changes, we ask whether intermediate updates tend to proceed in the same direction as the final outcome.
We hypothesize that for reasoning traces leading to accurate solutions, the sequence of updates should mostly advance toward the final representation.

Formally, this is assessed by comparing each update vector $v^{(n)}_l$ with the drift vector $u_l$.
The cosine similarity $\frac{\langle v^{(n)}_l, u_l\rangle}{||v^{(n)}_l||_2 ||u_l||_2}$ measures the angle between the two, indicating whether each local update proceeds in the same general direction as the overall displacement.
Averaging across segments and layers yields a single score:
\[
\textsc{AlignedChange} \;=\; \frac{1}{L}\sum_{l \in L}
\frac{1}{N-1}\sum_{n=2}^{N}
\frac{\langle v^{(n)}_l,\,u_l\rangle}{\|v^{(n)}_l\|_2\,\|u_l\|_2}.
\]

Higher values suggest that intermediate updates are aligned with the overall progress toward the final state, while lower values indicate they are inconsistent or even opposed to it.

\section{Experimental Setup}
\subsection{Baselines}
For baselines, we compare with two alternative approaches: (1) \emph{Cross-Layer Signals}, which summarize representational changes \emph{across layers} within a segment, and (2) \emph{Output Distribution Measures}, which estimate confidence from the token distribution at the final answer.

\noindent\textbf{Cross-Layer Signals:}
Previous work has shown that changes across layers can be predictive of answer accuracy in CoT. Following \cite{wang2024latent}, for each reasoning segment $n$, we compute the mean \emph{magnitude} and \emph{angle} of layer-to-layer changes and then average over segments:
\noindent
\begin{minipage}[t]{0.45\linewidth}
\[
\textsc{LayerMag}^{(n)} =
\frac{1}{L} \sum_{l=2}^{L}
\frac{\| \tilde{h}_{l} - \tilde{h}_{l-1} \|_2}{\| \tilde{h}_L - \tilde{h}_1 \|_2};
\]
\end{minipage}
\begin{minipage}[t]{0.49\linewidth}
\[
\textsc{LayerAng}^{(n)} =
\frac{1}{L} \sum_{l=2}^{L}
\frac{\arccos(\cos(\tilde{h}_l, \tilde{h}_{l-1}))}{\arccos(\cos(\tilde{h}_L, \tilde{h}_1))}
\]
\end{minipage}

\noindent\textbf{Output Distribution Measures:}
Output distribution–based measures are commonly used as estimates of model confidence \citep{yona2022useful, kadavath2022language, manakul-etal-2023-selfcheckgpt}, and can be used as proxies for final answer reliability.
To compare against these metrics, we elicit the final answer post reasoning trace end using prompts of the form [$\dots\{trace\_end\}$ \texttt{Final Answer:}], and examine the probability distribution over the token that follows.
Based on findings from \cite{yona2022useful}, we considered three best performing output distribution measures:
(i) \textbf{Logit Margin}: the difference between the top-2 token logits; (ii) \textbf{Entropy}: the entropy of the token distribution; (iii) \textbf{Perplexity}: the inverse probability of the model’s top-ranked token.

\subsection{Models and Datasets}
We evaluate three open-source reasoning models: Deepseek-R1-Distill-Qwen-14B (\textsc{R1-D}) \citep{guo2025deepseek}, Phi-4 Reasoning-Plus (\textsc{Phi4R+}) \citep{abdin2025phi}, and Qwen3-14B (\textsc{Qwen3}) \citep{yang2025qwen3}.
Our study tests our \method{} signals across three distinct reasoning domains: (i) \textit{Scientific}, measured using the \textit{GPQA Diamond} benchmark, which comprises 198 graduate-level multiple-choice questions in biology, chemistry, and physics \citep{rein2024gpqa}; (ii) \textit{Mathematical}, evaluated on \textit{AIME 2025}, a 30-problem set from the American Invitational Mathematics Examination \citep{AIME2025}; (iii) \textit{Algorithmic}, assessed with a stratified subsample ($n=180$) of the \textit{TSP} benchmark, consisting of path-optimization problems across varying levels of difficulty (graphs of 6 to 13 nodes) \citep{GeoMeterData_nphard_tsp1}.

\section{Results}

\begin{figure}
  \includegraphics[width=\textwidth]{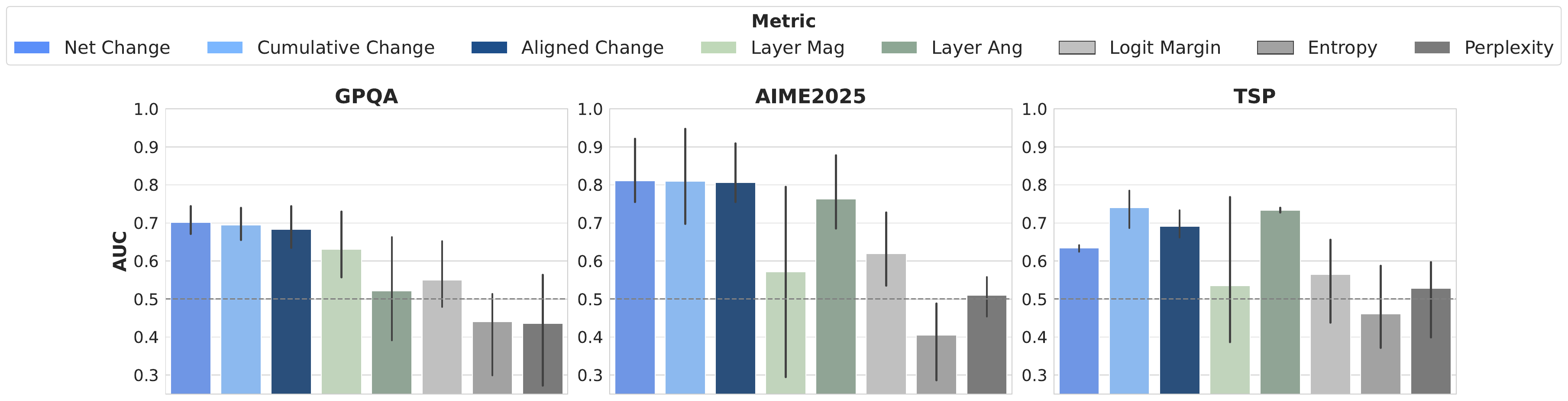}
  \vspace{-0.7cm}
  \caption{ROC-AUC for distinguishing correct from incorrect predictions using \method{} (LT) and baseline metrics. Higher values indicate better discriminative power. For comparability, Cumulative Change was sign-reversed. LT signals consistently achieve above chance (dashed line) and more reliable discrimination than baseline metrics. Error bars denote variability across models.
  }
  \label{metrics_auc}
  \vspace{-0.3cm}
\end{figure}

\subsection{\methodfull{} signals are predictive of solution accuracy}

To assess whether \method{} signals predict the correctness of the solution, we evaluate their discriminative power using the area under the ROC curve (AUC). 
For each problem, we generate five independent reasoning traces with their corresponding final answers and compute a \method{} score for each trace based on the hidden states of its intermediate reasoning tokens.
We then compute ROC-AUC with respect to accuracy by sweeping a decision threshold over the scores, which captures how well the signals distinguish correct from incorrect solutions.

As shown in Figure~\ref{metrics_auc}, \method{} signals significantly distinguish between reasoning traces that lead to accurate versus inaccurate answers.
\emph{Across datasets, the ROC-AUCs of our three \method{} signals remain consistently above chance, demonstrating robust predictive power} (Net Change mean ROC-AUC $= 0.71 \pm 0.09$; Cumulative Change $= 0.74 \pm 0.09$; Aligned Change $= 0.73 \pm 0.08$). 
In contrast, the cross-layer magnitude and angle signals are less reliable and vary substantially across models and reasoning domains (Cross-Layer Magnitude Change $= 0.58 \pm 0.17$; Cross-Layer Angle Change $=0.67 \pm 0.14$).
Output-distribution–based metrics are significantly weaker and less consistent, with performance often close to or below chance level (Logit Margin $= 0.59 \pm 0.10$; Entropy $= 0.44 \pm 0.10$, Perplexity $= 0.49 \pm 0.12$). 
In summary, our results show that for models that produce long intermediate traces, signals that capture the temporal evolution in latent space are stronger and more robust predictors of solution accuracy than cross-layer geometry or output-distribution-based confidence measures (see Appendix \ref{appendix_predictivity} for ROC-AUC scores for each model-dataset combination).

\begin{wrapfigure}{r}{0.65\textwidth}
\vspace{-1.2em} 
\centering
\includegraphics[width=0.64\textwidth,trim=20 0 20 0,clip]{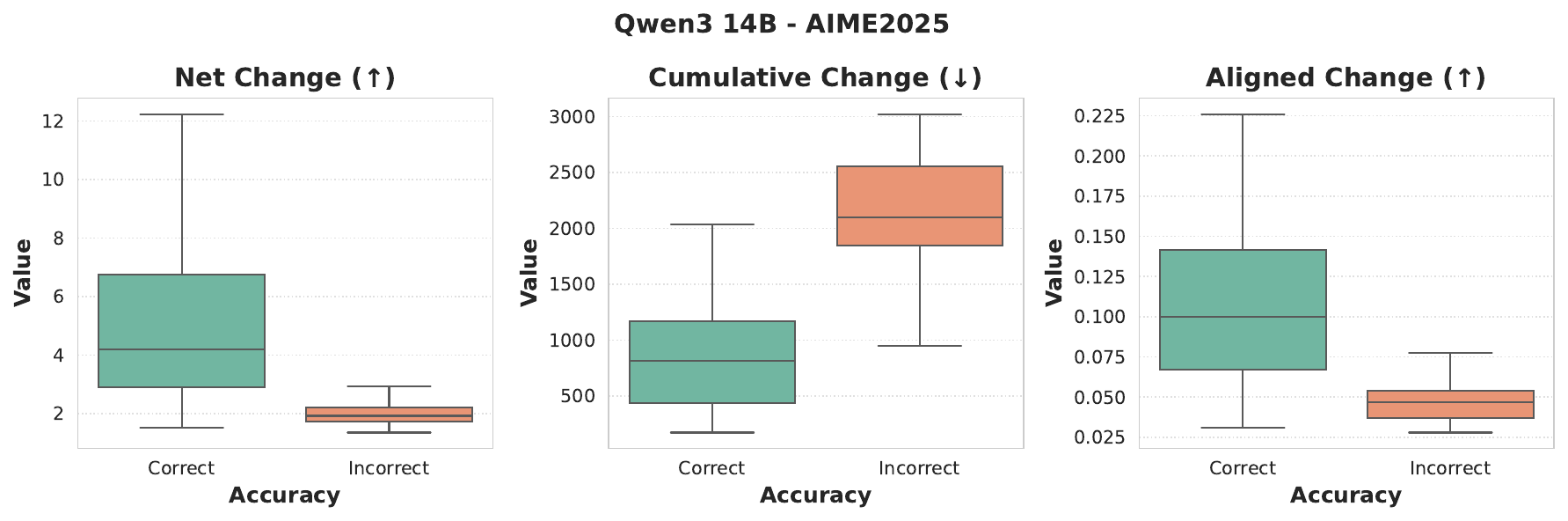}
\vspace{-1em} 
\caption{Latent-Trajectory signal distributions by accuracy for Qwen3-14B on the AIME 2025 dataset. 
Correct traces show larger Net/Aligned Change and smaller Cumulative Change than incorrect ones. This indicates that correct reasoning corresponds to larger, more directed representational shifts, while incorrect reasoning involves more wandering and less aligned trajectories.}
\label{distribution_plot}
\vspace{-1.2em}
\end{wrapfigure}

We found that Cumulative Change was negatively correlated with accuracy (Spearman's $r = -.38$), which indicates that traces that traverse greater total distance in representation space tend to be less likely to produce correct answers.
This finding mechanistically grounds prior behavioral observations that long but highly varying reasoning traces are associated with lower accuracy \citep{balachandran2025inference, shojaee2025illusion}.
Net and Aligned Change show positive associations with accuracy (Net Change $=.28$; Aligned Change $= .32$). 
Larger overall representational change from the initial to the final hidden state is therefore linked to better performance, and representational updates that progress more directly toward the final state show an even stronger association. 
Figure~\ref{distribution_plot} shows the distributions of the three trajectory metrics for Qwen3 on AIME2025. 
The distribution of values further supports our findings: successful trajectories cover greater distances in latent space, advance more directly toward the final state at intermediate steps, and involve less path deviations.
Equivalent plots for each model and dataset are in Appendix~\ref{appendix_dist_plots}, including plots of layer-wise values for each \method{} signal.

\subsection{\methodfull{} signals improve efficiency and reliability of multi-sample inference}

Building on our previous finding that \method{} signals strongly predict solution accuracy, we now investigate whether they can guide more accurate and efficient scaling strategies in multi-sample inference systems.
Previous work demonstrates that generating multiple answers and aggregating them through self-consistency improves both accuracy and reliability in language models \citep{wang2023selfconsistency, Kang2025SelfCertaintyBoN}. 
In practice, majority voting (MV) has become the default approach for recent releases of reasoning models \citep{abdin2025phi, guo2025deepseek}, since a single inference pass is rarely sufficient for robust performance, especially in applications or agentic settings \citep{Besta2025TopologiesReasoningTPAMI}.
This robustness, however, comes with increased inference costs, particularly for reasoning models, where long chains of thought lead token usage to grow by an order of magnitude with each additional sample. 
Here, we examine whether \method{}-based selection can preserve the benefits of MV while reducing sample and token budget.

\begin{wrapfigure}{r}{0.55\textwidth}
\vspace{-1.5em} 
\centering
\includegraphics[width=0.54\textwidth,trim=20 0 20 0,clip]{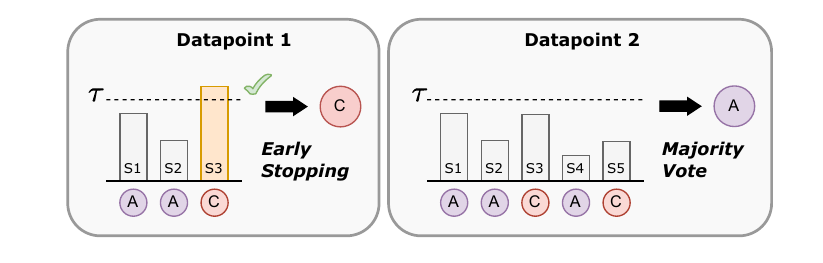}
\vspace{-1.5em} 
\caption{Candidate solutions for a problem are evaluated sequentially. If a solution’s signal value exceeds $\tau$, it is immediately accepted as the final prediction. If no solution crosses $\tau$, the final answer is chosen via MV.}
\label{framework}
\end{wrapfigure}

\noindent \textbf{Experiment Setup:} 
We generate multiple samples sequentially from the model and use \method{} signals to decide online whether the current trace is likely correct and should be used as the final answer, or whether additional samples are needed.
Once a signal exceeds a calibrated threshold, we accept the solution early and stop sampling.
If no samples cross the threshold after at most $k$ attempts, we fall back to MV over the collected candidates (see Figure \ref{framework}). 
This allows datapoints with strong internal signals to be resolved quickly with fewer samples, while datapoints with weaker signals rely on the robustness of aggregation.
We set $k=5$ and repeat this procedure independently for each signal. 

We compare our approach against two sample aggregation baselines: (i) MV, and (ii) shortest-answer selection, which chooses out of the sampled answers the candidate with the fewest tokens, motivated by recent findings that shorter completions are strong signals of accuracy \citep{hassid2025shorter, shrivastava2025sample, marjanovic2025thoughtology}. 

We select decision thresholds $\tau$ using a cross-validation approach (see Appendix \ref{appendix_CV_threhsold} for details).
On the calibration set, we construct candidate thresholds from quantiles of the metric values observed among incorrect solutions, such that each threshold corresponds to a fixed proportion of errors exceeding the cutoff.
For each candidate, we simulated the full decision rule on the calibration subset, accepting a solution early when the signal crosses the threshold and otherwise aggregating with MV.
The best-calibrated threshold is then evaluated on the remaining data.
We report accuracy as the fraction of problems solved correctly, and efficiency in terms of (i) the average number of samples required, and (ii) the proportion of reasoning tokens consumed relative to running the full inference procedure with five samples. 
Reported results are averaged over the splits.

In addition to exploring each metric separately, we built a \textbf{Combined \method{} score} from a weighted sum of the \method{} values, where signals that are more strongly associated with accuracy on the calibration set contributed more to the final score (Appendix \ref{appendix_combined_score} reports details on its construction).

\noindent\textbf{Results:} As shown in Table \ref{tab:threshold_experiments}, \method{} signals improve both efficiency and accuracy relative to MV.
On GPQA, R1-D gains on average $2\%$, Qwen3 remains stable, and Phi4R+ maintains competitive accuracy. 
On AIME2025, improvements are more pronounced with $4\%$ for R1-D, $2\%$ for Phi4R+, and a substantial $12\%$ for Qwen3. 
On TSP, all models benefit, with consistent gains of $1$--$3\%$. 
These results show that \method{} thresholds not only preserve correctness across settings, but often deliver meaningful boosts by identifying correct reasoning paths even when the majority of solutions are incorrect.
Efficiency gains are considerably larger. 
On R1-D, \method{} signals reduce the average number of tokens required to match or outperform MV by \(50\)–\(66\%\) across datasets, Qwen3 achieves reductions of about $50$-$55\%$, and Phi4R+ reduces samples by $30$--$35\%$. 
Across all settings, the Shortest@5 baseline reduces accuracy by an average of $1.4\%$, showing that length alone is an unreliable proxy for correctness. 
In Appendix \ref{appendix_inference_experiments}, we further report that \method{} signals enable early answer selection for $>$85\% of data points on average across datasets, and that accuracy within this subset consistently exceeds the baselines, confirming that \method{} signals concentrate probability mass on more reliable solutions.

Overall, \emph{Latent-Trajectory-guided selection preserves, and often improves, the reliability of majority-vote aggregation while substantially reducing inference costs}.
The Combined \method{} score is frequently competitive with the best individual signal and, in most cases, cuts token usage by at least half.
This makes it a practical and effective choice when applied to different models and datasets.
At an aggregate level, compared to MV@5, \method{} strategies exhibit:
(i) \textit{Sample savings}: Number of sampled answers is reduced on average by \textbf{58\%} ($32$--$76$\%); (ii) \textit{Token savings}: As a consequence of sample savings, token usage (and thereby inference cost) is reduced on average by \textbf{48\%} ($14$--$70$\%); (iii) \textit{Accuracy improvement}: Accuracy increases on average by \textbf{2.64\%} ($-1.4$--$14.10$\%).

\definecolor{BetterGreen}{RGB}{0,128,0}   
\definecolor{BetterRed}{RGB}{178,34,34}   
\definecolor{ForestGreen}{RGB}{34,139,34} 
\definecolor{BrickRed}{RGB}{178,34,34}
\newcommand{\posc}[1]{\begingroup\color{ForestGreen}#1\endgroup}
\newcommand{\negc}[1]{\begingroup\color{BrickRed}#1\endgroup}

\begin{table}[H]
\centering
\footnotesize
\vspace{-1.5em}
\newcommand{\NA}{\multicolumn{1}{c}{--}}
\caption{Accuracy and efficiency with Latent-Trajectory  (LT) signals. Baselines are MV@5 (majority vote across 5 samples) and Shortest@5 (shortest of 5 samples). 
Accuracy is reported as a percentage, with parentheses indicating the change relative to MV@5. For efficiency, we report the average number of samples required per datapoint, with parentheses showing the percentage reduction in total token usage relative to MV. \underline{\textbf{Bold}} and \textbf{Bold} denote the best and second-best results within each group. \smallcheck{} marks cases where the average number of samples was reduced by more than half. 
}
\label{tab:threshold_experiments}
\resizebox{\linewidth}{!}{%
\begin{tabular}{
  @{}llllllll@{}
}
\toprule
& & \multicolumn{2}{c}{\textbf{GPQA}} &
    \multicolumn{2}{c}{\textbf{AIME2025}} &
    \multicolumn{2}{c}{\textbf{TSP}} \\
\cmidrule(lr){3-4}\cmidrule(lr){5-6}\cmidrule(lr){7-8}
\textbf{Model} & \textbf{Strategy} &
\multicolumn{1}{c}{\makecell{Acc. \\(avg \% / $\Delta$Acc)}} & 
\multicolumn{1}{c}{\makecell{Samples\\(avg / $\Delta$Tok \%)}} &
\multicolumn{1}{c}{\makecell{Acc.\\(avg \% / $\Delta$Acc)}} & 
\multicolumn{1}{c}{\makecell{Samples\\(avg / $\Delta$Tok \%)}} &
\multicolumn{1}{c}{\makecell{Acc.\\(avg \% / $\Delta$Acc)}} & 
\multicolumn{1}{c}{\makecell{Samples\\(avg / $\Delta$Tok \%)}} \\
\midrule
\multirow{6}{*}{R1-D}
  & MV@5 & 59.90 & 5.00  & 56.67 & 5.00  & 27.50 & 5.00  \\
  & Shortest@5 & 60.91 \posc{(+1.0)} & 5.00 \text{ (0)}  & 50.00 \negc{(-6.7)} & 5.00 \text{ (0)} & 28.75 \posc{(+1.3)} & 5.00 \text{ (0)} \\
  & \multicolumn{1}{>{\columncolor{gray!10}}l}{LT -- Net} & \multicolumn{1}{>{\columncolor{gray!10}}l}{61.10 \posc{(+1.2)}} & \multicolumn{1}{>{\columncolor{gray!10}}l}{\textbf{1.69} \posc{(+53.9)} \smallcheck} & \multicolumn{1}{>{\columncolor{gray!10}}l}{\underline{\textbf{61.90}} \posc{(+5.2)}} & \multicolumn{1}{>{\columncolor{gray!10}}l}{\underline{\textbf{1.22}} \posc{(+68.7)} \smallcheck} & \multicolumn{1}{>{\columncolor{gray!10}}l}{28.60 \posc{(+1.1)}} & \multicolumn{1}{>{\columncolor{gray!10}}l}{\underline{\textbf{1.43}} \posc{(+70.6)} \smallcheck} \\
  & \multicolumn{1}{>{\columncolor{gray!10}}l}{LT -- Cumulative} & \multicolumn{1}{>{\columncolor{gray!10}}l}{\underline{\textbf{62.10}} \posc{(+2.2)}} & \multicolumn{1}{>{\columncolor{gray!10}}l}{1.88 \posc{(+48.1)} \smallcheck} & \multicolumn{1}{>{\columncolor{gray!10}}l}{58.70 \posc{(+2.0)}} & \multicolumn{1}{>{\columncolor{gray!10}}l}{2.56 \posc{(+29.9)}} & \multicolumn{1}{>{\columncolor{gray!10}}l}{\underline{\textbf{30.90}} \posc{(+3.4)}} & \multicolumn{1}{>{\columncolor{gray!10}}l}{\textbf{1.61} \posc{(+66.4)} \smallcheck} \\
  & \multicolumn{1}{>{\columncolor{gray!10}}l}{LT -- Aligned} & \multicolumn{1}{>{\columncolor{gray!10}}l}{61.10 \posc{(+1.2)}} & \multicolumn{1}{>{\columncolor{gray!10}}l}{\underline{\textbf{1.58}} \posc{(+57.0)} \smallcheck} & \multicolumn{1}{>{\columncolor{gray!10}}l}{\textbf{60.30} \posc{(+3.6)}} & \multicolumn{1}{>{\columncolor{gray!10}}l}{\textbf{1.43} \posc{(+61.3)} \smallcheck} & \multicolumn{1}{>{\columncolor{gray!10}}l}{29.50 \posc{(+2.0)}} & \multicolumn{1}{>{\columncolor{gray!10}}l}{2.08 \posc{(+57.2)} \smallcheck} \\
  & \multicolumn{1}{>{\columncolor{gray!10}}l}{LT -- Combined} & \multicolumn{1}{>{\columncolor{gray!10}}l}{\textbf{61.80} \posc{(+1.9)}} & \multicolumn{1}{>{\columncolor{gray!10}}l}{1.89 \posc{(+47.3)} \smallcheck} & \multicolumn{1}{>{\columncolor{gray!10}}l}{\underline{\textbf{61.90}} \posc{(+5.2)}} & \multicolumn{1}{>{\columncolor{gray!10}}l}{2.06 \posc{(+43.9)} \smallcheck} & \multicolumn{1}{>{\columncolor{gray!10}}l}{\textbf{30.10} \posc{(+2.6)}} & \multicolumn{1}{>{\columncolor{gray!10}}l}{\textbf{\underline{1.43}} \posc{(+70.3)} \smallcheck} \\
\midrule
\multirow{6}{*}{Phi4R+}
  & MV@5 & \underline{\textbf{70.20}} & 5.00  & 80.00 & 5.00  & 41.25 & 5.00  \\
  & Shortest@5 & 69.19 \negc{(-1.0)} & 5.00 \text{ (0)} & 70.00 \negc{(-10.0)} & 5.00 \text{ (0)} & 38.75 \negc{(-2.5)} & 5.00 \text{ (0)} \\
  & \multicolumn{1}{>{\columncolor{gray!10}}l}{LT -- Net} & \multicolumn{1}{>{\columncolor{gray!10}}l}{68.80 \negc{(-1.4)}} & \multicolumn{1}{>{\columncolor{gray!10}}l}{\textbf{\underline{2.97}} \posc{(+20.2)}} & \multicolumn{1}{>{\columncolor{gray!10}}l}{79.40 \negc{(-0.6)}} & \multicolumn{1}{>{\columncolor{gray!10}}l}{\textbf{\underline{2.19}} \posc{(+41.1)} \smallcheck} & \multicolumn{1}{>{\columncolor{gray!10}}l}{42.30 \posc{(+1.1)}} & \multicolumn{1}{>{\columncolor{gray!10}}l}{\textbf{\underline{1.59}} \posc{(+67.2)} \smallcheck} \\
  & \multicolumn{1}{>{\columncolor{gray!10}}l}{LT -- Cumulative} & \multicolumn{1}{>{\columncolor{gray!10}}l}{\textbf{69.60} \negc{(-0.6)}} & \multicolumn{1}{>{\columncolor{gray!10}}l}{\textbf{2.99} \posc{(+18.9)}} & \multicolumn{1}{>{\columncolor{gray!10}}l}{81.00 \posc{(+1.0)}} & \multicolumn{1}{>{\columncolor{gray!10}}l}{\textbf{2.43} \posc{(+32.1)} \smallcheck} & \multicolumn{1}{>{\columncolor{gray!10}}l}{\textbf{\underline{44.40}} \posc{(+3.1)}} & \multicolumn{1}{>{\columncolor{gray!10}}l}{2.63 \posc{(+42.2)}} \\
  & \multicolumn{1}{>{\columncolor{gray!10}}l}{LT -- Aligned} & \multicolumn{1}{>{\columncolor{gray!10}}l}{\textbf{69.60} \negc{(-0.6)}} & \multicolumn{1}{>{\columncolor{gray!10}}l}{3.40 \posc{(+14.5)}} & \multicolumn{1}{>{\columncolor{gray!10}}l}{\textbf{82.50} \posc{(+2.5)}} & \multicolumn{1}{>{\columncolor{gray!10}}l}{2.51 \posc{(+30.8)}} & \multicolumn{1}{>{\columncolor{gray!10}}l}{\textbf{44.10} \posc{(+2.9)}} & \multicolumn{1}{>{\columncolor{gray!10}}l}{\textbf{1.96} \posc{(+58.7)} \smallcheck} \\
  & \multicolumn{1}{>{\columncolor{gray!10}}l}{LT -- Combined} & \multicolumn{1}{>{\columncolor{gray!10}}l}{\textbf{69.60} \negc{(-0.6)}} & \multicolumn{1}{>{\columncolor{gray!10}}l}{3.28 \posc{(+16.3)}} & \multicolumn{1}{>{\columncolor{gray!10}}l}{\textbf{\underline{82.60}} \posc{(+2.6)}} & \multicolumn{1}{>{\columncolor{gray!10}}l}{2.54 \posc{(+28.7)}} & \multicolumn{1}{>{\columncolor{gray!10}}l}{43.80 \posc{(+2.6)}} & \multicolumn{1}{>{\columncolor{gray!10}}l}{2.30 \posc{(+50.5)} \smallcheck} \\
\midrule
\multirow{6}{*}{Qwen3}
  & MV@5 & 63.96 & 5.00  & 70.00 & 5.00 & 36.25 & 5.00  \\
  & Shortest@5 & \textbf{\underline{64.47}} \posc{(+0.5)} & 5.00 \text{ (0)} & 80.00 \posc{(+10.0)} & 5.00 \text{ (0)} & 30.63 \negc{(-5.6)} & 5.00 \text{ (0)} \\
  & \multicolumn{1}{>{\columncolor{gray!10}}l}{LT -- Net} & \multicolumn{1}{>{\columncolor{gray!10}}l}{63.70 \negc{(-0.3)}} & \multicolumn{1}{>{\columncolor{gray!10}}l}{\textbf{\underline{1.42}} \posc{(+63.9)} \smallcheck} & \multicolumn{1}{>{\columncolor{gray!10}}l}{79.40 \posc{(+9.4)}} & \multicolumn{1}{>{\columncolor{gray!10}}l}{1.60 \posc{(+57.3)} \smallcheck} & \multicolumn{1}{>{\columncolor{gray!10}}l}{35.40 \negc{(-0.9)}} & \multicolumn{1}{>{\columncolor{gray!10}}l}{3.18 \posc{(+34.0)}} \\
  & \multicolumn{1}{>{\columncolor{gray!10}}l}{LT -- Cumulative} & \multicolumn{1}{>{\columncolor{gray!10}}l}{63.30 \negc{(-0.7)}} & \multicolumn{1}{>{\columncolor{gray!10}}l}{2.25 \posc{(+41.3)} \smallcheck} & \multicolumn{1}{>{\columncolor{gray!10}}l}{\textbf{\underline{84.10}} \posc{(+14.1)}} & \multicolumn{1}{>{\columncolor{gray!10}}l}{2.03 \posc{(+43.2)} \smallcheck} & \multicolumn{1}{>{\columncolor{gray!10}}l}{\textbf{36.30} \posc{(+0.1)}} & \multicolumn{1}{>{\columncolor{gray!10}}l}{\textbf{\underline{1.64}} \posc{(+65.5)} \smallcheck} \\
  & \multicolumn{1}{>{\columncolor{gray!10}}l}{LT -- Aligned} & \multicolumn{1}{>{\columncolor{gray!10}}l}{\textbf{64.20} \posc{(+0.2)}} & \multicolumn{1}{>{\columncolor{gray!10}}l}{1.75 \posc{(+52.0)} \smallcheck} & \multicolumn{1}{>{\columncolor{gray!10}}l}{\textbf{80.90} \posc{(+10.9)}} & \multicolumn{1}{>{\columncolor{gray!10}}l}{\textbf{1.59} \posc{(+58.2)} \smallcheck} & \multicolumn{1}{>{\columncolor{gray!10}}l}{\textbf{\underline{37.80}} \posc{(+1.6)}} & \multicolumn{1}{>{\columncolor{gray!10}}l}{\textbf{2.08} \posc{(+56.4)} \smallcheck} \\
  & \multicolumn{1}{>{\columncolor{gray!10}}l}{LT -- Combined} & \multicolumn{1}{>{\columncolor{gray!10}}l}{63.70 \negc{(-0.3)}} & \multicolumn{1}{>{\columncolor{gray!10}}l}{\textbf{1.70} \posc{(+53.4)} \smallcheck} & \multicolumn{1}{>{\columncolor{gray!10}}l}{\textbf{80.90} \posc{(+10.9)}} & \multicolumn{1}{>{\columncolor{gray!10}}l}{\textbf{\underline{1.49}} \posc{(+60.3)} \smallcheck} & \multicolumn{1}{>{\columncolor{gray!10}}l}{36.00 \negc{(-0.3)}} & \multicolumn{1}{>{\columncolor{gray!10}}l}{2.46 \posc{(+48.4)} \smallcheck} \\
\bottomrule
\end{tabular}
}
\end{table}

\subsection{\methodfull{} signals enable early selection of high-quality traces}

\paragraph{Experiment Setup:} 
While the previous section focused on using \method{} signals for end-of-trace answer selection, we now ask whether these signals can also identify higher-quality trajectories early in the reasoning process.
To investigate this, we run a step-wise early-exit evaluation. 
We evaluate signals on \emph{partial} traces taken at 500-token intervals up to the full trace.
At each checkpoint, we recompute Net Change and Cumulative Change\footnote{As Aligned Change compares the direction of each segment with respect to the last segment, it is inconsistent when applied earlier in the trace.} using only the tokens available so far. 
For each partial trace, the most recent segment is used as the final segment.
We then compute the ROC-AUC of each signal at every checkpoint, revealing how predictive power evolves as the trace unfolds, and whether prediction of solution correctness is possible without observing the full trajectory.

To investigate whether these early signals in the reasoning trace can be leveraged during inference, we implement an early path selection policy when sampling multiple generations in parallel. 
At 2k tokens, we compute the \method{} signals on the partial traces and use them as features for a lightweight random forest classifier trained to predict correctness. 
The classifier selects a single candidate trajectory to continue, while the other four paths are terminated. 
The chosen trace is decoded to completion, and we report both the accuracy of this early path selection policy and the proportion of tokens saved compared to running all five trajectories to the end.

\begin{figure}[h]  \includegraphics[width=\linewidth]{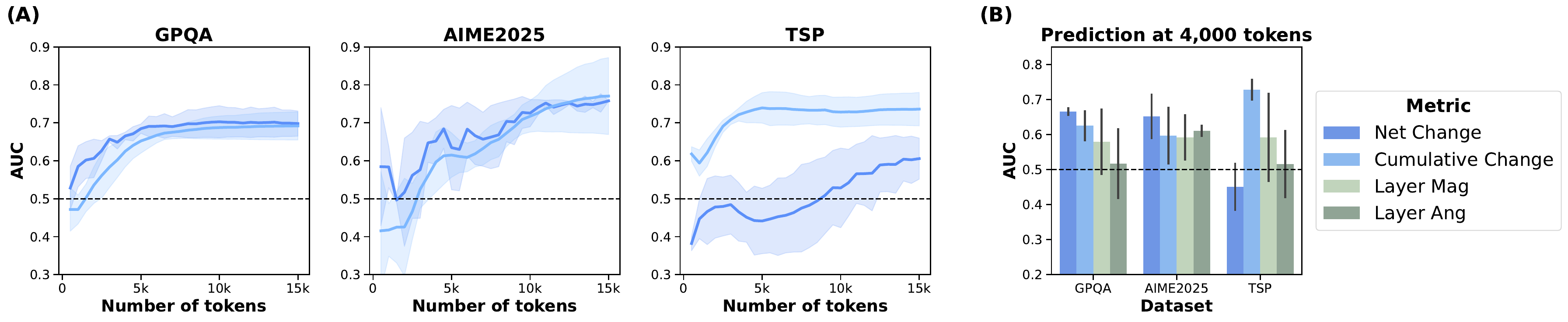}
\vspace{-0.6cm}
  \caption{
\textbf{(A)} Predictive performance (ROC-AUC) of Net and Cumulative Change signals as a function of the number of tokens, across datasets. 
Shaded regions represent variation across models. 
\textbf{(B)} Comparison of predictive performance at 4k tokens. Error bands indicate variation across models. 
Performance of \method{} signals rises early well above the 0.5 baseline.
}
  \label{early_exit}
\end{figure}

\paragraph{Results:} As Figure \ref{early_exit} shows, Net and Cumulative change provide early in the trace predictive signals well above chance, with ROC-AUC generally increasing as additional tokens are observed. 
ROC-AUC values above $.6$ can be obtained within the first 4k tokens, with Net Change being a better predictor than Cumulative Change early in the trace for GPQA and AIME2025.
This pattern, however, reverses for TSP, 
where Cumulative Change is significantly more predictive than Net Change throughout the early and mid-trace.

At inference-time, early selection of high-quality paths using \method{} substantially reduces computational cost while maintaining or improving accuracy (see Table~\ref{tab:early_exit_experiments}). 
Across models and datasets, accuracy remains highly competitive with MV@5: R1-D achieves a $6.7\%$ gain on AIME2025 with only negligible differences on GPQA and TSP; 
Phi4R+ improves accuracy by $2$--$4\%$ across datasets;
and Qwen3 yields gains of $2$--$3\%$. 
Efficiency improvements are even larger, with average token usage reduced by $50$--$65\%$ for R1-D and Qwen3, and by $70\%$ for Phi4R+.
On aggregate, we observe (i) an average increase of \textbf{2.1\%}; and (ii) an average token reduction of \textbf{61\%}.
These results demonstrate that \emph{\method{} enables early selection of high-quality reasoning paths, allocating compute to the most promising generations and achieving accuracy comparable to majority voting at less than half the inference cost}.

\begin{table}[H]
\centering
\footnotesize
\vspace{-1.5em}
\caption{Evaluation of early path selection (at 2k tokens) using \method{} (LT) signals. Accuracy (\%) and Saved Tokens (\%) with $\Delta$ relative to Majority Vote (Maj@5).}
\label{tab:early_exit_experiments}
\resizebox{\linewidth}{!}{%
\begin{tabular}{
  @{}l l
  cc cc cc
  @{}
}
\toprule
& & \multicolumn{2}{c}{\textbf{GPQA}} &
    \multicolumn{2}{c}{\textbf{AIME2025}} &
    \multicolumn{2}{c}{\textbf{TSP}} \\
\cmidrule(lr){3-4}\cmidrule(lr){5-6}\cmidrule(lr){7-8}
\textbf{Model} & \textbf{Strategy} &
\multicolumn{1}{c}{\makecell{Accuracy\\(\% / $\Delta$ \%)}} &
\multicolumn{1}{c}{\makecell{Saved Tokens\\($\Delta$ \%)}} &
\multicolumn{1}{c}{\makecell{Accuracy\\(\% / $\Delta$ \%)}} &
\multicolumn{1}{c}{\makecell{Saved Tokens\\($\Delta$ \%)}} &
\multicolumn{1}{c}{\makecell{Accuracy\\(\% / $\Delta$ \%)}} &
\multicolumn{1}{c}{\makecell{Saved Tokens\\($\Delta$ \%)}} \\
\midrule
R1-D & Maj@5 & 59.90 & - & 56.67 & - & 27.50 & - \\
& \cellcolor{gray!10} LT & \cellcolor{gray!10} 59.39 \negc{(-0.5)} & \cellcolor{gray!10} \posc{+48.9} & \cellcolor{gray!10} 63.33 \posc{(+6.7)} & \cellcolor{gray!10} \posc{+50.1} & \cellcolor{gray!10} 26.25 \negc{(-1.3)} & \cellcolor{gray!10} \posc{+62.5} \\
\midrule
Phi4R+ & Maj@5 & 70.20 & - & 80.00 & - & 41.25 & - \\
& \cellcolor{gray!10} LT & \cellcolor{gray!10} 72.22 \posc{(+2.0)} & \cellcolor{gray!10} \posc{+64.7} & \cellcolor{gray!10} 83.33 \posc{(+3.3)} & \cellcolor{gray!10} \posc{+67.3} & \cellcolor{gray!10} 45.63 \posc{(+4.4)} & \cellcolor{gray!10} \posc{+71.7} \\
\midrule
Qwen3 & Maj@5 & 63.96 & - & 70.00 & - & 36.25 & - \\
& \cellcolor{gray!10} LT & \cellcolor{gray!10} 66.50 \posc{(+2.5)} & \cellcolor{gray!10} \posc{+51.0} & \cellcolor{gray!10} 73.33 \posc{(+3.3)} & \cellcolor{gray!10} \posc{+69.1} & \cellcolor{gray!10} 38.13 \posc{(+1.9)} & \cellcolor{gray!10} \posc{+65.7} \\
\bottomrule
\end{tabular}
}
\end{table}

\section{Conclusions}
Our work introduced a family of \method{} signals that capture the temporal evolution of reasoning traces within a model's latent space. 
Across multiple reasoning domains and models, \method{} metrics predict final-answer correctness significantly above chance and outperform other internal and output-distribution-based baselines. 
We further demonstrated their utility in practical test-time policies. 
In inference-scaling experiments, using these signals for answer selection or early path selection reduced token usage and often improved accuracy with respect to strong baselines such as majority vote.
Our efficiency gains address two complementary sources of inefficiency: 
(i) reducing the number of samples required for reliable reasoning, 
and (ii) shortening individual trajectories by detecting early answers of higher quality.
The approach is model-agnostic, simple to calibrate, and compatible with existing sampling strategies.  
In addition to practical benefits, our results shed light on the structure of reasoning in latent space, revealing how trajectories unfold during inference and what distinguishes successful from unsuccessful reasoning paths.

There are several opportunities for future work. 
While we show the real-world utility of these signals at inference time, trajectory-level signals could also provide actionable guidance for fine-tuning and calibration, with the potential to guide models toward more reliable reasoning trajectories.
In addition, our study introduced lightweight techniques for metric aggregation and threshold selection. 
An exciting direction for future work is to explore learned classifiers or ensembles to further boost the informativeness of the signals.

\clearpage


\subsubsection*{Acknowledgments}
We thank Xavier Fernandez, Subbarao Kambhampati, Vibhav Vineet, 
Lingjiao Chen, Tyler LaBonte, Jiwan Chung, Wanjia Zhao, Erfan Shayegani, Ieva Bagdonaviciute, Ahmed Awadallah for their invaluable support and feedback throughout this project.

\bibliography{iclr2026_conference}
\bibliographystyle{iclr2026_conference}

\clearpage

\appendix

\section{Predictivity of \methodfull{} Signals}
\label{appendix_predictivity}
In Table \ref{tab:metrics_eval} we provide the ROC-AUC and Spearman's $r$ values with accuracy for each Latent-Trajectory (LT) and baseline metric computed per model-dataset pair.

\begin{table}[H]
\centering
\footnotesize
\caption{ROC-AUC and correlation (Spearman's $r$) with accuracy for each model-dataset pair.}
\label{tab:metrics_eval}
\resizebox{\linewidth}{!}{%
\begin{tabular}{@{}llcccccc@{}}
\toprule
& & \multicolumn{2}{c}{\textbf{GPQA}} &
    \multicolumn{2}{c}{\textbf{AIME2025}} &
    \multicolumn{2}{c}{\textbf{TSP}} \\
\cmidrule(lr){3-4}\cmidrule(lr){5-6}\cmidrule(lr){7-8}
\textbf{Model} & \textbf{Metric} &
\multicolumn{1}{c}{\makecell{AUC}} & \multicolumn{1}{c}{\makecell{Corr.}} &
\multicolumn{1}{c}{\makecell{AUC}} & \multicolumn{1}{c}{\makecell{Corr.}} &
\multicolumn{1}{c}{\makecell{AUC}} & \multicolumn{1}{c}{\makecell{Corr.}} \\
\midrule
\multirow{8}{*}{R1-D}
  & \multicolumn{1}{l}{\cellcolor{gray!20}Net Change} & \cellcolor{gray!20}.688 & \cellcolor{gray!20}.320 & \cellcolor{gray!20}.757 & \cellcolor{gray!20}.433 & \cellcolor{gray!20}.641 & \cellcolor{gray!20}.223 \\
  & \multicolumn{1}{l}{\cellcolor{gray!20}Cumulative Change} & \cellcolor{gray!20}.690 & \cellcolor{gray!20}-.323 & \cellcolor{gray!20}.697 & \cellcolor{gray!20}-.333 & \cellcolor{gray!20}.687 & \cellcolor{gray!20}-.294 \\
  & \multicolumn{1}{l}{\cellcolor{gray!20}Aligned Change} & \cellcolor{gray!20}.670 & \cellcolor{gray!20}.288 & \cellcolor{gray!20}.755 & \cellcolor{gray!20}.430 & \cellcolor{gray!20}.662 & \cellcolor{gray!20}.255 \\
  & Layer Magnitude         & .606 & .180 & .795 &  .497 & .449 & -.080 \\
  & Layer Angle         & .392 & .184 & .685 & -.313 & .740 & -.378 \\
  & Logit Margin      & .554 & .092 & .597 & .163 & .656 & .246 \\
  & Entropy           & .510 & .016 & .488 & -.020 & .588 & .138 \\
  & Perplexity        & .656 & .266 & .558 & .097 & .597 & .153 \\
\midrule
\multirow{8}{*}{Phi4R+}
  & \multicolumn{1}{l}{\cellcolor{gray!20}Net Change} & \cellcolor{gray!20}.744 & \cellcolor{gray!20}.391 & \cellcolor{gray!20}.755 & \cellcolor{gray!20}.366 & \cellcolor{gray!20}.625 & \cellcolor{gray!20}.433 \\
  & \multicolumn{1}{l}{\cellcolor{gray!20}Cumulative Change} & \cellcolor{gray!20}.740 & \cellcolor{gray!20}-.384 & \cellcolor{gray!20}.786 & \cellcolor{gray!20}-.410 & \cellcolor{gray!20}.785 & \cellcolor{gray!20}-.333 \\
  & \multicolumn{1}{l}{\cellcolor{gray!20}Aligned Change} & \cellcolor{gray!20}.744 & \cellcolor{gray!20}.391 & \cellcolor{gray!20}.755 & \cellcolor{gray!20}.366 & \cellcolor{gray!20}.733 & \cellcolor{gray!20}.430 \\
  & Layer Magnitude         & .730 & .368 & .625 & .180 & .768 & .497 \\
  & Layer Angle         & .663 & -.260 & .727 & -.325 & .733 & -.312 \\
  & Logit Margin      & .652 & .243 & .535 & .050 & .438 & .163 \\
  & Entropy           & .230 & -.321 & .440 & -.086 & .426 & -.020 \\
  & Perplexity        & .369 & -.194 & .517 & .024 & .399 & .097 \\
\midrule
\multirow{8}{*}{Qwen3}
  & \multicolumn{1}{l}{\cellcolor{gray!20}Net Change} & \cellcolor{gray!20}.671 & \cellcolor{gray!20}.286 & \cellcolor{gray!20}.921 & \cellcolor{gray!20}.651 & \cellcolor{gray!20}.637 & \cellcolor{gray!20}.229 \\
  & \multicolumn{1}{l}{\cellcolor{gray!20}Cumulative Change} & \cellcolor{gray!20}.655 & \cellcolor{gray!20}-.259 & \cellcolor{gray!20}.947 & \cellcolor{gray!20}-.691 & \cellcolor{gray!20}.748 & \cellcolor{gray!20}-.414 \\
  & \multicolumn{1}{l}{\cellcolor{gray!20}Aligned Change} & \cellcolor{gray!20}.635 & \cellcolor{gray!20}.225 & \cellcolor{gray!20}.909 & \cellcolor{gray!20}.632 & \cellcolor{gray!20}.679 & \cellcolor{gray!20}.300 \\
  & Layer Magnitude         & .557 & .095 & .295 & -.317 & .387 & -.189 \\
  & Layer Angle         & .509 & -.015 & .878 & -.584 & .727 & .380 \\
  & Logit Margin      & .444 & -.094 & .728 & .351 & .602 & .170 \\
  & Entropy           & .513 & .022 & .286 & -.324 & .371 & -.213 \\
  & Perplexity        & .272 & -.380 & .454 & -.071 & .591 & .151 \\
\bottomrule
\end{tabular}
}
\end{table}

\clearpage

\section{Latent-Trajectory Signals}
\label{appendix_dist_plots}

To investigate how \method{} dynamics relate to model performance, we compare distributions of our three representational signals—Net Change, Cumulative Change, and Aligned Change—conditioned on whether a model’s final answer was correct or incorrect. 
Figures 4–6 present box plots of these metrics across datasets and model families. 
These visualizations allow us to assess whether systematic differences in \method{} signals are associated with answer correctness, and whether such effects are consistent across evaluation settings.

\begin{figure}[H]
\includegraphics[width=\linewidth]{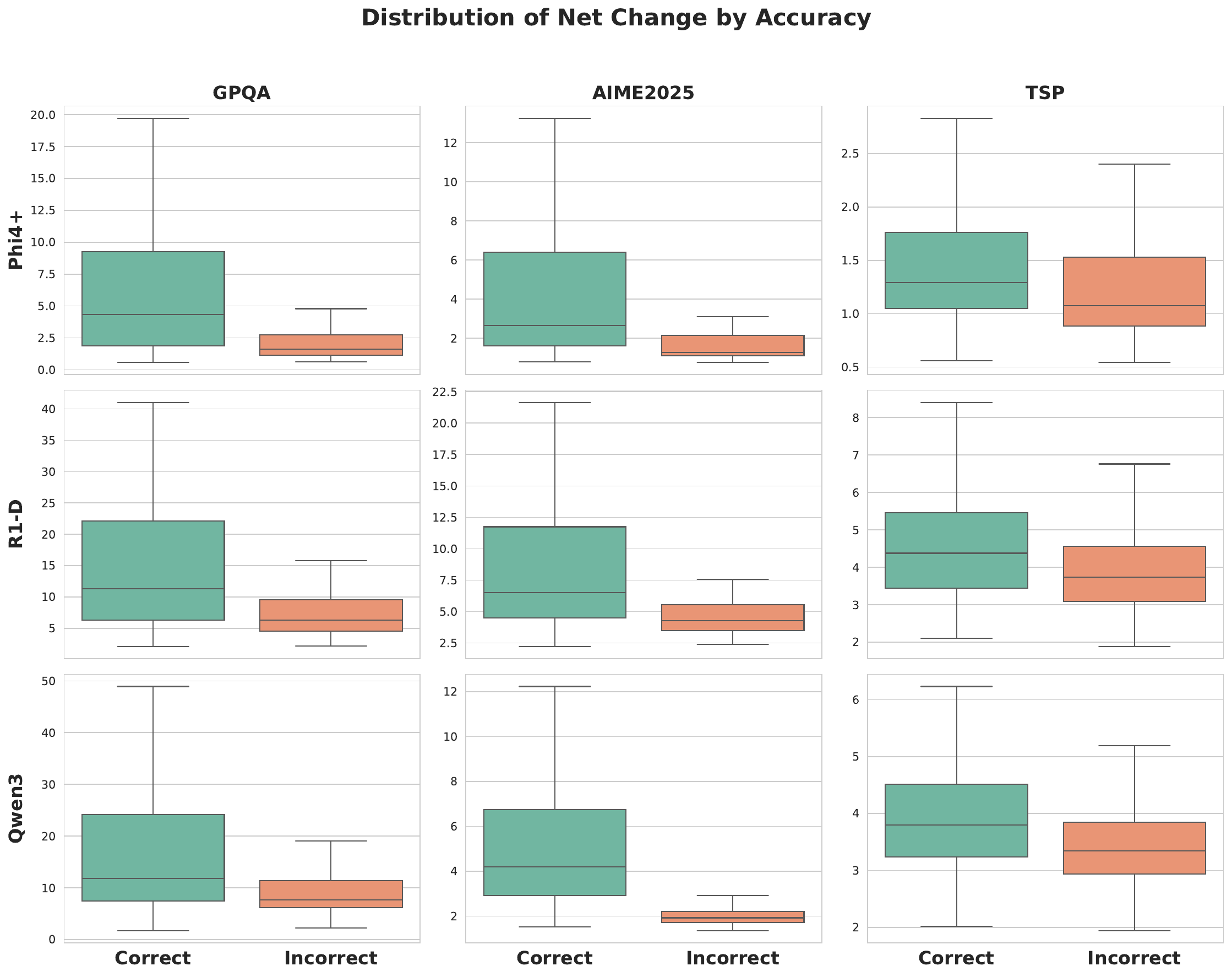}
  \caption{Distribution of Net Change by accuracy. Values are generally higher for correct than for incorrect responses, suggesting that successful reasoning is associated with overall larger representational drifts, which may be a sign of deeper reasoning.}
  \label{individual_metrics_plots_net}
\end{figure}

\begin{figure}[H]
\includegraphics[width=\linewidth]{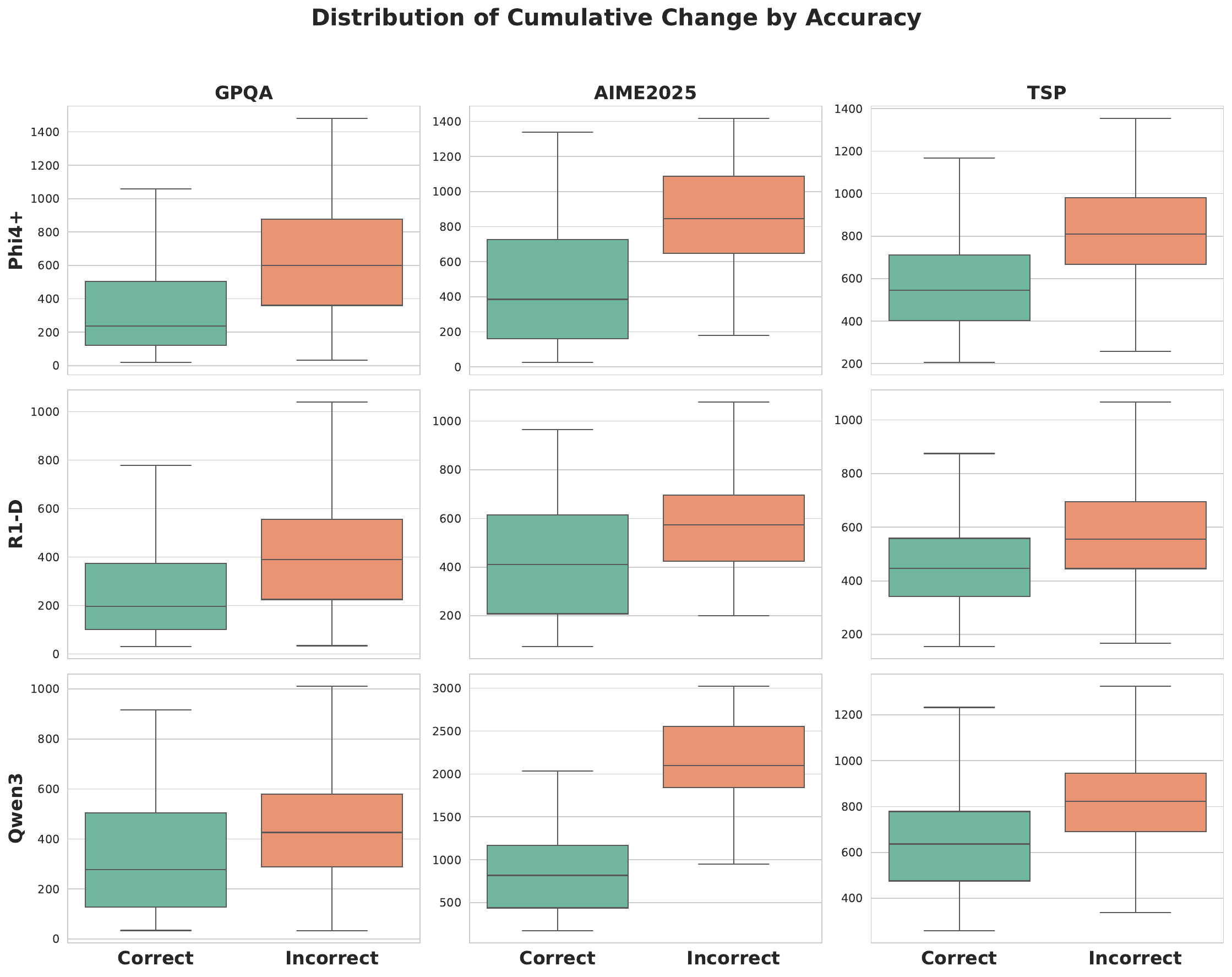}
  \caption{Distribution of Cumulative Change by accuracy. Values are often larger for incorrect responses, indicating that when models answer incorrectly, their latent trajectories tend to involve more movement through representational space, potentially reflecting less stable reasoning. }
  \label{individual_metrics_plots_cumulative}
\end{figure}

\begin{figure}[H]
\includegraphics[width=\linewidth]{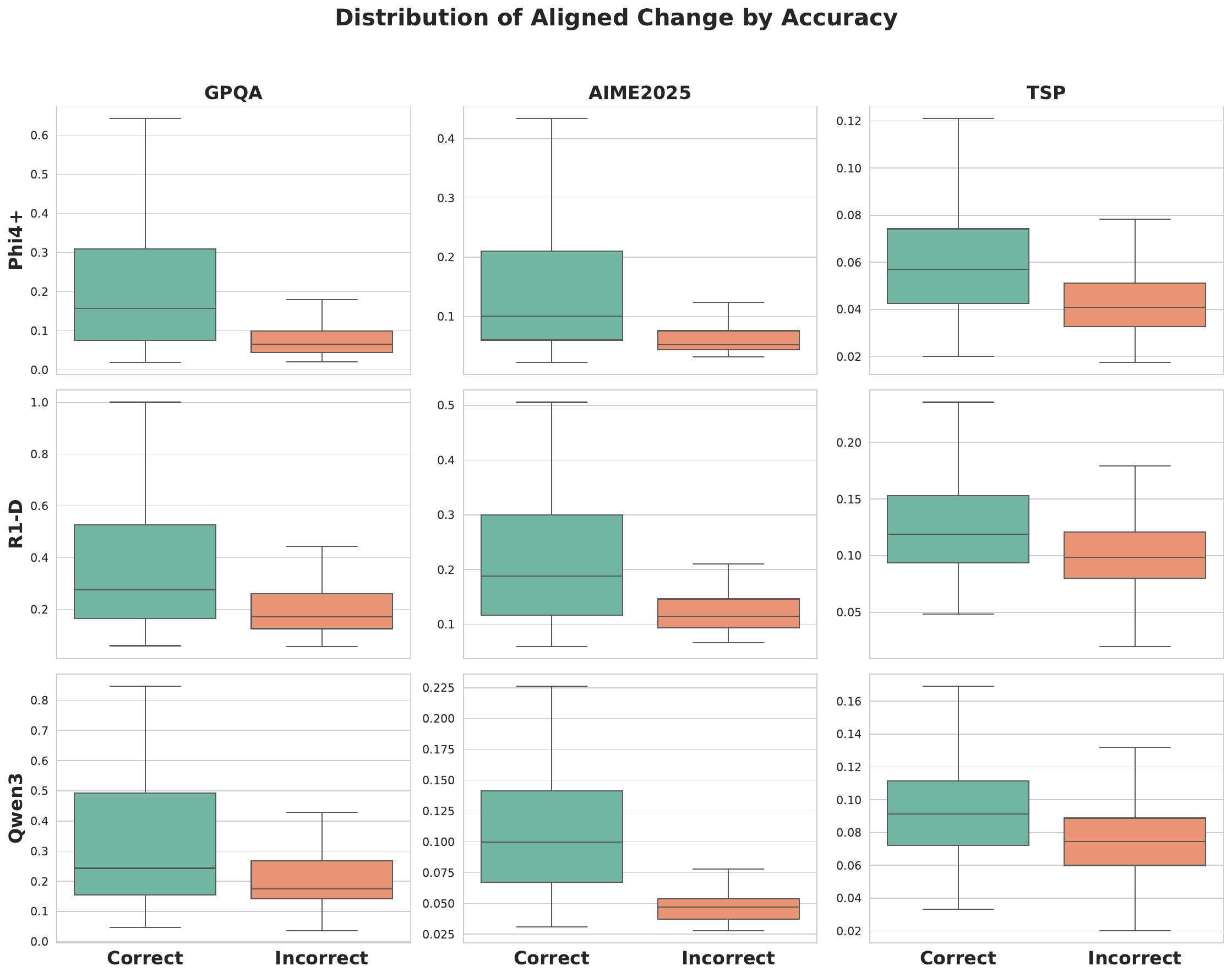}
  \caption{Distribution of Aligned Change by accuracy. Values are higher for correct responses, implying that effective reasoning involves intermediate representational updates that advance more directly towards the final state.}
  \label{individual_metrics_plots_aligned}
\end{figure}

For each of the signals, consistent patterns emerge. 
Net Change values (Figure \ref{individual_metrics_plots_net}) are generally higher for correct responses than for incorrect responses, suggesting that successful reasoning is associated with overall larger representational drifts.
In contrast, Cumulative Change values (Figure \ref{individual_metrics_plots_cumulative}) are often larger for incorrect responses, indicating that when models answer incorrectly, their latent trajectories tend to involve more movement through representational space, potentially reflecting less stable reasoning. 
Finally, Aligned Change values (Figure \ref{individual_metrics_plots_aligned}) are higher for correct responses, implying that effective reasoning requires updates that advance more directly towards the final state.

Taken together, these results suggest that correct predictions are characterized by a larger overall representational shift, accompanied by trajectories that are more directionally consistent, whereas incorrect predictions tend to involve longer, less aligned paths through latent space, reflecting noisier and less stable reasoning trajectories. This pattern holds across models and datasets, indicating that \method{} signals provide complementary and reliable signals of reasoning quality.

\begin{figure}[H]
\includegraphics[width=\linewidth]{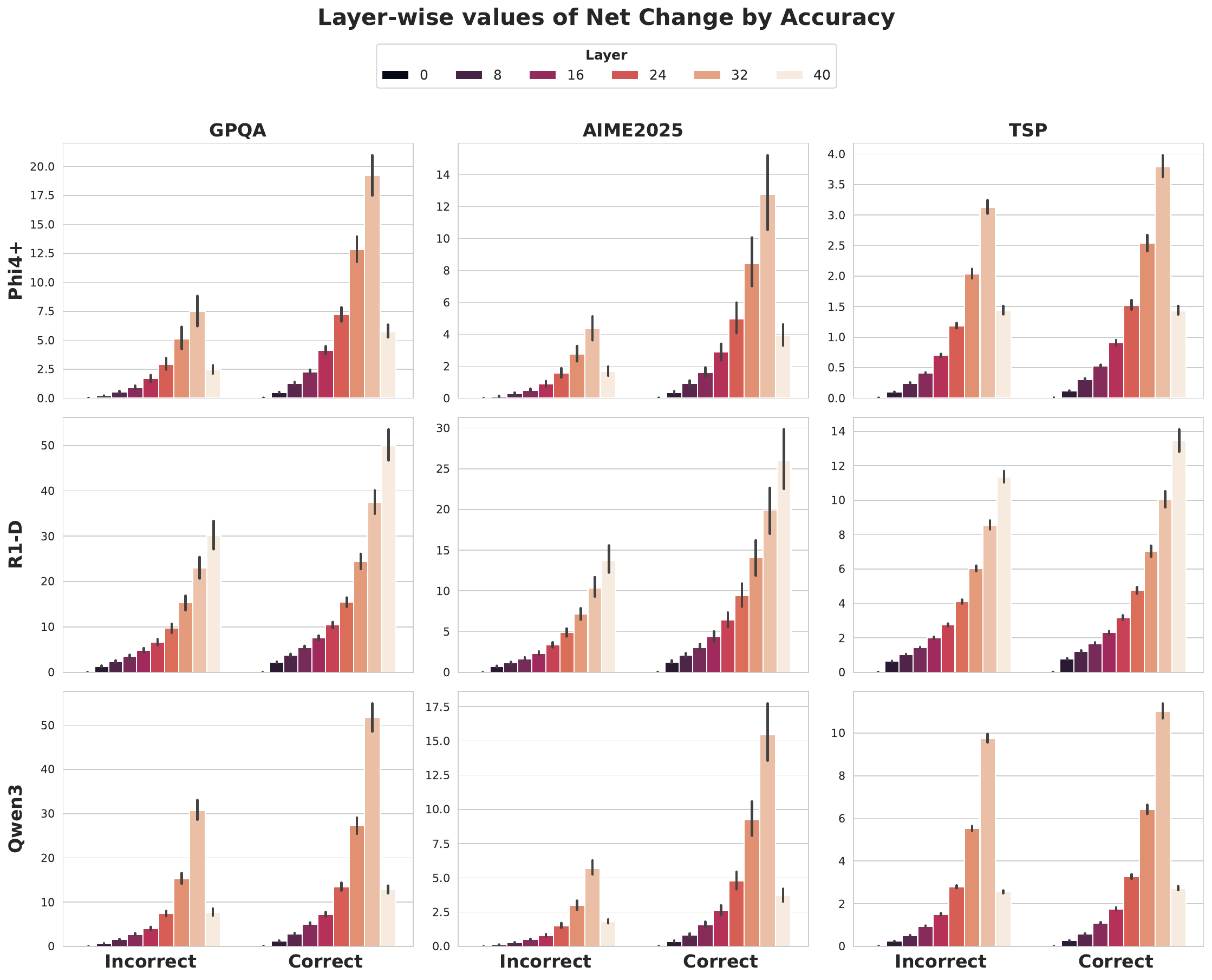}
  \caption{Layer-wise Net Change values for correct vs. incorrect reasoning traces across models and benchmarks. Correct trajectories generally show larger representational shifts across layers compared to incorrect ones, indicating that stronger changes are associated with successful reasoning.}
  \label{layerwise_net}
\end{figure}

\begin{figure}[H]
\includegraphics[width=\linewidth]{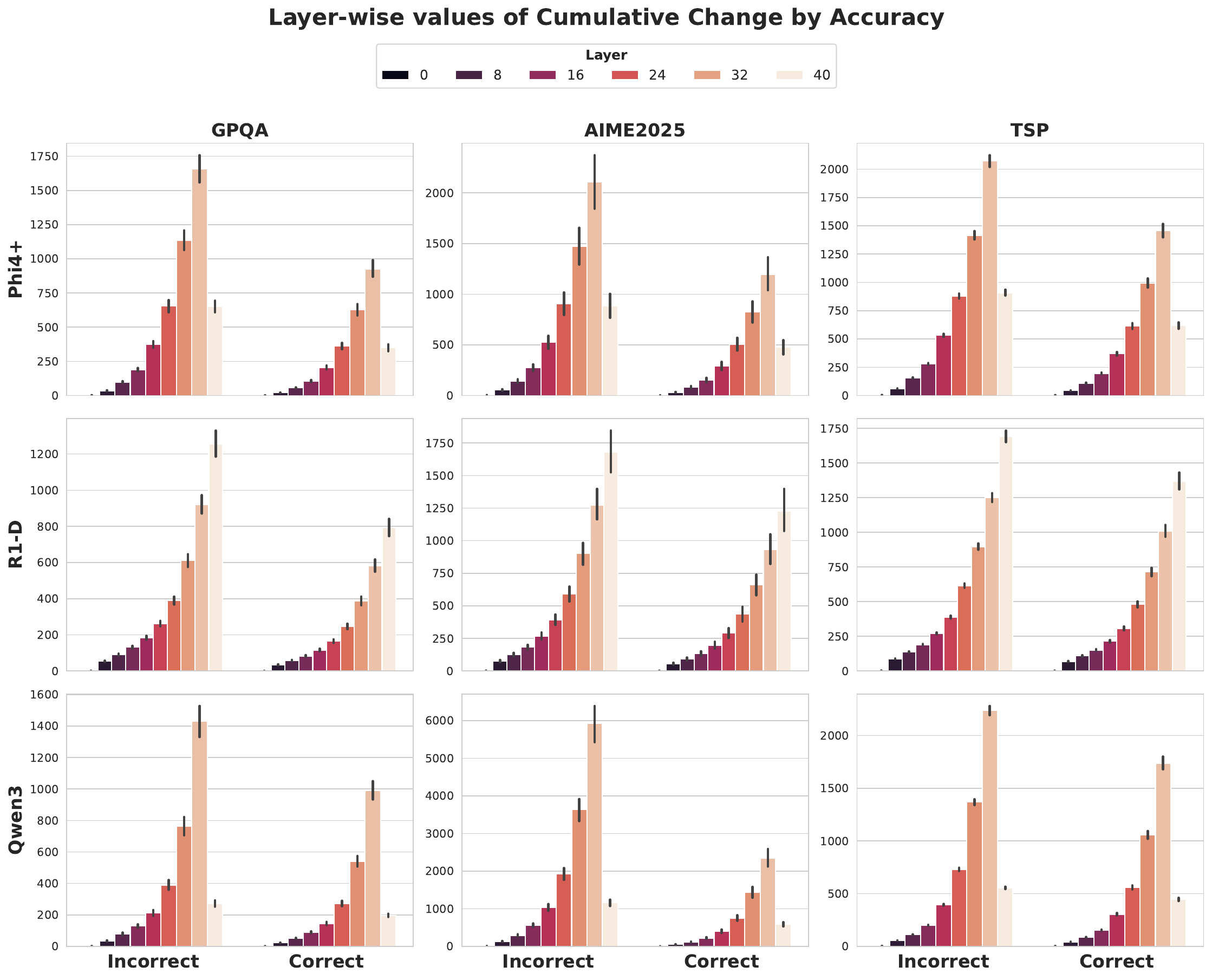}
  \caption{Layer-wise Cumulative Change values for correct vs. incorrect reasoning traces across models and benchmarks. Incorrect trajectories accumulate substantially larger representational shifts across layers than correct ones, indicating that unsuccessful reasoning is associated with less stable and more circuitous latent dynamics, indicating that correct traces take more direct paths towards the final solution.}
  \label{layerwise_path}
\end{figure}

\begin{figure}[H]
\includegraphics[width=\linewidth]{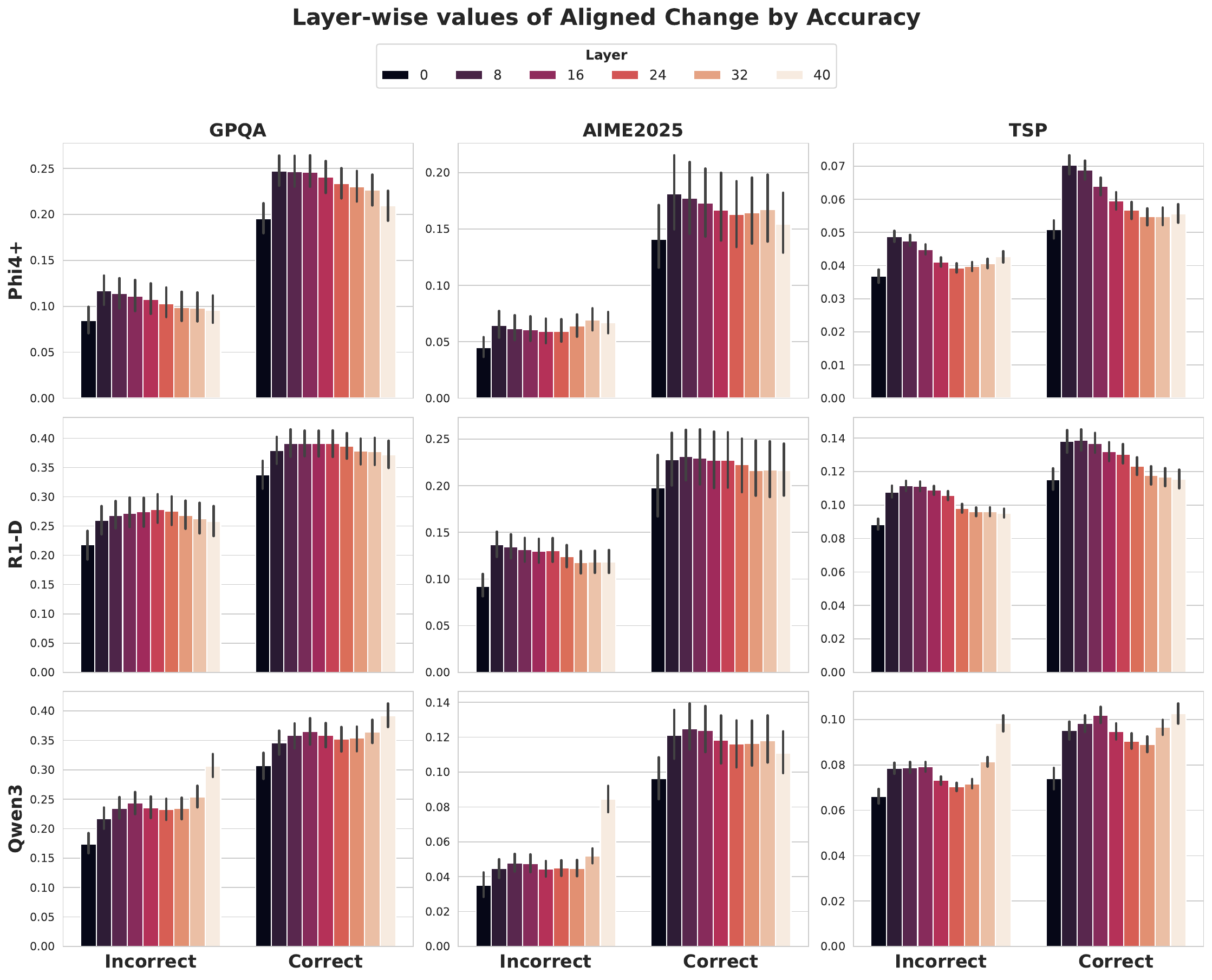}
  \caption{Layer-wise Aligned Change values for correct vs. incorrect reasoning traces across models and benchmarks. Correct trajectories consistently exhibit stronger alignment across layers compared to incorrect ones.}
  \label{layerwise_aligned}
\end{figure}

\clearpage
We also provide the average layer-wise values of each \method{} signal in Figure \ref{layerwise_net}, \ref{layerwise_path}, and \ref{layerwise_aligned}.
Each subplot compares how internal changes evolve by layer.

\clearpage

\clearpage

\clearpage

\section{Latent-Trajectory Signals for Inference-Time Scaling}
\label{appendix_inference_experiments}

In the main text, we evaluated efficiency and reliability when applying \method{} thresholds in a sequential inference procedure. Here, we provide additional analyses focusing on the subset of samples that exceeded the thresholds. This allows us to directly quantify (i) the accuracy of solutions accepted early and (ii) the proportion of datapoints where the \method{} decision rule was triggered.

Table \ref{tab:extended_lt_experiments} reports accuracy and coverage for above-threshold samples across models and datasets. Accuracy here refers only to the subset of candidate solutions whose \method{} score surpassed the calibrated threshold, while the “Datapoints” column indicates the fraction of evaluation datapoints where an early stop occurred. 
As expected, above-threshold samples are consistently more accurate than the overall average, often approaching ceiling performance for stricter thresholds. At the same time, the coverage varies: some learned thresholds are more lenient, allowing the rule to apply to a larger fraction of datapoints, while others are stricter, isolating a smaller but more reliable subset.

\begin{table}[H]
\centering
\footnotesize
\caption{Above threshold evaluation across models and datasets with \method{} (LT) strategies. Accuracy (\%) of samples above threshold, and percentage of Datapoints where LT decision rule was applied.}
\label{tab:extended_lt_experiments}
\resizebox{\linewidth}{!}{%
\begin{tabular}{
  @{}l l
  cc cc cc
  @{}
}
\toprule
& & \multicolumn{2}{c}{\textbf{GPQA}} &
    \multicolumn{2}{c}{\textbf{AIME2025}} &
    \multicolumn{2}{c}{\textbf{TSP}} \\
\cmidrule(lr){3-4}\cmidrule(lr){5-6}\cmidrule(lr){7-8}
\textbf{Model} & \textbf{Strategy} &
\multicolumn{1}{c}{\makecell{ Acc.\\(\%)}} & \multicolumn{1}{c}{\makecell{Datapoints\\(\%)}}  &
\multicolumn{1}{c}{\makecell{ Acc.\\(\%)}}  & \multicolumn{1}{c}{\makecell{Datapoints\\(\%)}} &
\multicolumn{1}{c}{\makecell{ Acc.\\(\%)}}  & \multicolumn{1}{c}{\makecell{Datapoints\\(\%)}} \\
\midrule
\multirow{4}{*}{R1-D}
  & LT -- Net & 64.60 & 88.2 & 68.73 & 90.5 & 28.73 & 99.4 \\
  & LT -- Cumulative & 67.57 & 81.4 & 75.63 & 54.0 & 32.13 & 96.4 \\
  & LT -- Aligned & 63.77 & 92.0 & 73.00 & 82.5 & 31.60 & 93.4 \\
  & LT -- Combined & 66.50 & 81.2 & 78.53 & 65.1 & 30.53 & 98.5 \\
\midrule
\multirow{4}{*}{Phi4R+}
  & LT -- Net & 84.37 & 60.0 & 82.20 & 80.9 & 42.03 & 97.0 \\
  & LT -- Cumulative & 84.87 & 57.1 & 89.53 & 74.6 & 56.20 & 74.7 \\
  & LT -- Aligned & 88.10 & 44.6 & 89.33 & 71.4 & 49.73 & 88.7 \\
  & LT -- Combined & 86.30 & 48.0 & 91.10 & 69.8 & 51.53 & 81.5 \\
\midrule
\multirow{4}{*}{Qwen3}
  & LT -- Net & 65.17 & 95.6 & 87.13 & 92.1 & 45.87 & 61.6 \\
  & LT -- Cumulative & 69.47 & 74.1 & 96.50 & 85.7 & 37.90 & 95.8 \\
  & LT -- Aligned & 64.73 & 87.7 & 85.37 & 95.2 & 40.50 & 87.5 \\
  & LT -- Combined & 65.53 & 88.2 & 85.37 & 95.2 & 47.43 & 74.4 \\
\bottomrule
\end{tabular}
}
\end{table}

\begin{figure}[h]
\includegraphics[width=\linewidth]{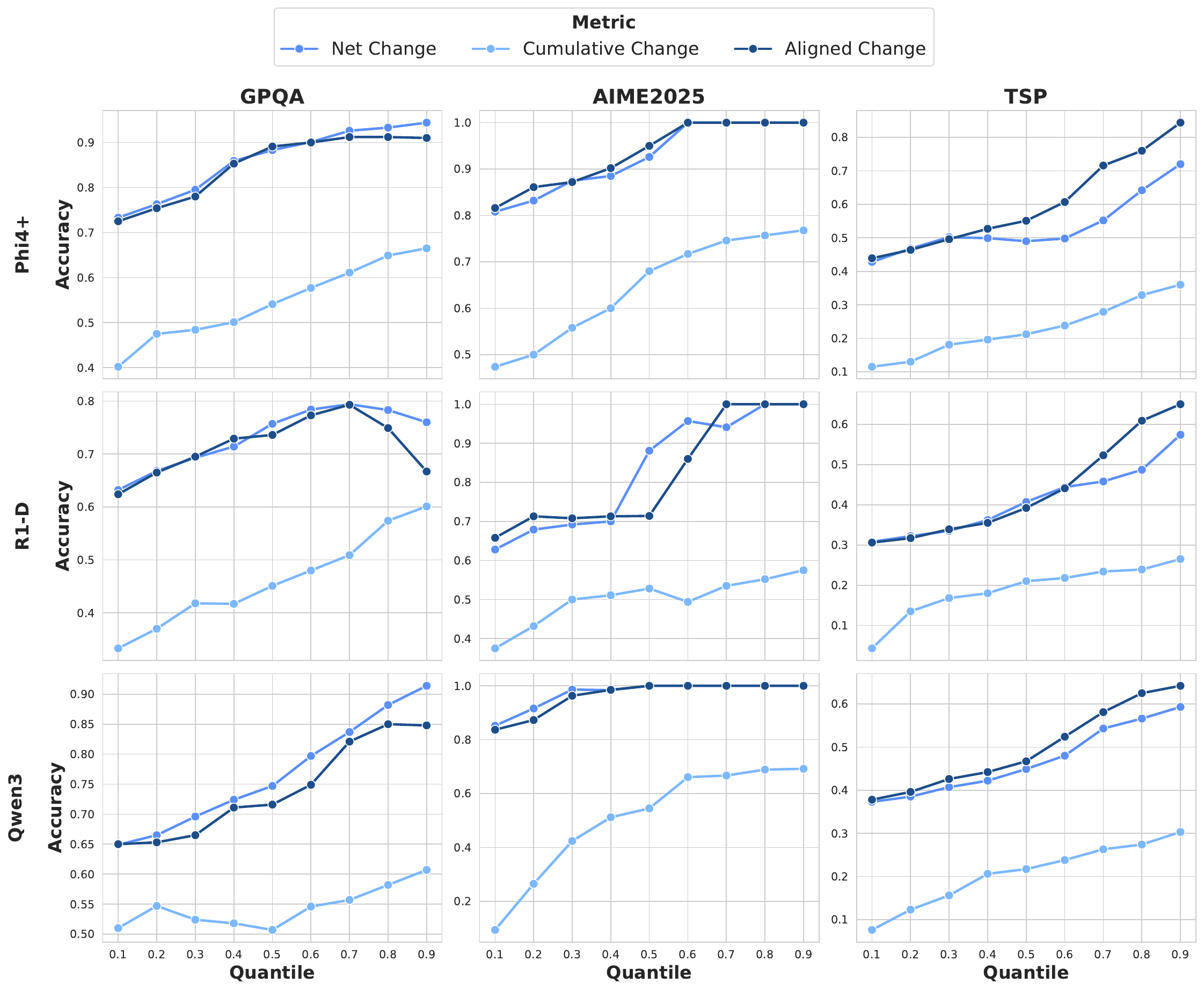}
  \caption{Accuracy of datapoints above thresholds defined over a range of quantiles. Given that Cumulative Change is negatively correlated with accuracy, we inverted the quantile selection (i.e., replacing $q$ with $1-q$) so that higher quantiles consistently correspond to higher expected accuracy. For all metrics, accuracy increases consistently with higher quantiles across datasets and models. This evidences that these metrics are predictive of answer quality. }
  \label{threshold_accuracy}
\end{figure}

To further illustrate this tradeoff, Figure \ref{threshold_accuracy} plots accuracy as a function of threshold quantiles. Higher quantiles consistently yield higher accuracy across all metrics, models, and datasets, indicating that \method{} signals reliably concentrate correct solutions in their upper ranges. In several cases, accuracy at the top quantiles approaches $100\%$, demonstrating that filtering by strong \method{} signals isolates highly reliable reasoning traces.

\clearpage

\clearpage

\section{Calibration procedure for threshold selection}
\label{appendix_CV_threhsold}

We use a three-fold shuffled cross-validation procedure. In each split, 30\% of the data is set aside for calibration, and the remaining 70\% is reserved for testing, with the same random seed applied across folds to ensure consistency.

\paragraph{Candidate thresholds.}
Within each calibration fold and for each metric, we focus on the subset of datapoints where the model’s answer was incorrect. 
If this subset contains fewer than 15 examples or if the metric has no valid values, we default to using the median value of the metric on the calibration set as the threshold. 
Otherwise, we construct a grid of candidate thresholds by taking the 20th through 99th percentiles of the metric values among the incorrect examples.

\paragraph{Evaluating a candidate threshold.}
For each threshold in this grid, calibration datapoints are divided into two groups. The first group consists of datapoints where at least one candidate solution exceeds the threshold. For these, we accept the first candidate that crosses the threshold and record its accuracy. The second group contains the remaining datapoints, which are resolved using majority vote across their candidates. The overall calibration accuracy for a given threshold is computed as the weighted average of the accuracies from these two groups, proportional to their sizes.

\paragraph{Selecting the threshold.}
We then rank thresholds by their overall calibration accuracy. The two best-performing thresholds are identified, and we set the final calibration threshold to their median.

\paragraph{Direction of comparison.}
For most metrics, higher values indicate stronger signals, so the threshold rule is applied as \({\rm metric}\ge t\). The Cumulative Change signals behave in the opposite direction, with smaller values being more predictive; here the rule is applied as \({\rm metric}\le t\).

\clearpage

\section{Combined Latent-Trajectory Score}
\label{appendix_combined_score}
In addition to exploring each metric separately, we built a \textbf{Combined \method{} score}.
For each dataset, we quantified the predictive utility of each signal by calculating its absolute Pearson correlation with accuracy on a 10\% calibration slice of the dataset.
We align directions so that larger values always indicate better performance (i.e. Cumulative Change was sign-inverted). 
We then normalize the correlations to obtain weights that sum to one, which yields an interpretable distribution of relative importance across metrics.
The combined \method{} score for each sampled solution is a weighted sum of the \method{} values, where signals that are more strongly associated with accuracy on the calibration set contributed more to the final score.
Table \ref{tab:combined-metric-weights} reports the weights.

\begin{table}[h]
\centering
\small
\caption{Metric weights for Combined Latent Space score.}
\label{tab:combined-metric-weights}
\setlength{\tabcolsep}{6pt}
\begin{tabular}{@{}llS[table-format=1.2]S[table-format=1.2]S[table-format=1.2]@{}}
\toprule
\textbf{Model} & \textbf{Dataset} &
\multicolumn{1}{c}{\makecell{Net\\Change}} &
\multicolumn{1}{c}{\makecell{Cumulative\\Change}} &
\multicolumn{1}{c}{\makecell{Aligned\\Change}} \\
\midrule
R1-D & GPQA & 0.35 & 0.40 & 0.25 \\
Phi4R+ & GPQA & 0.30 & 0.38 & 0.31 \\
Qwen3 & GPQA & 0.43 & 0.25 & 0.32 \\
\midrule
R1-D & AIME2025 & 0.31 & 0.35 & 0.34 \\
Phi4R+ & AIME2025 & 0.26 & 0.45 & 0.29 \\
Qwen3 & AIME2025 & 0.30 & 0.43 & 0.28 \\
\midrule
R1-D & TSP & 0.24 & 0.39 & 0.37 \\
Phi4R+ & TSP & 0.20 & 0.38 & 0.42 \\
Qwen3 & TSP & 0.19 & 0.43 & 0.37 \\
\bottomrule
\end{tabular}
\end{table}

\clearpage

\section{Representational Averaging}
\label{appendix_averaging}
To enhance the signal robustness and reduce the dimensionality of the reasoning trace, we partition it into non-overlapping \textbf{reasoning segments} of size \textbf{500}.
For each layer, we then average the token representations within each segment.
We found that this procedure preserves the overall trajectory of the reasoning process.

\begin{figure}[H]
\includegraphics[width=\linewidth]{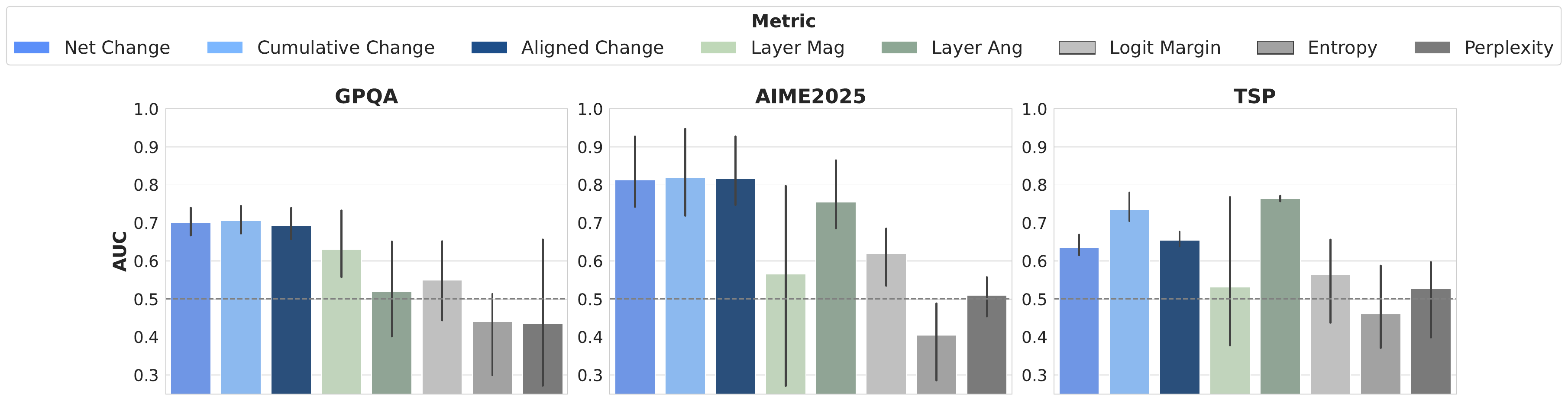}
  \caption{ROC-AUC for distinguishing correct from incorrect predictions using \method{} (LT) and baseline metrics with reasoning segments of \textbf{300} token length. Higher values indicate better discriminative power. For comparability, Cumulative Change was sign-reversed. LT signals consistently achieve above chance (dashed line) and more reliable discrimination than baseline metrics. Error bars denote variability across models.}
  \label{300_tokens}
\end{figure}

The choice of 500 tokens was guided by the average answer lengths across datasets. 
The dataset with the shortest responses still had an average of 5,000 tokens per answer. 
Setting the window to 500 tokens therefore ensures that, on average, we obtain at least 10 measurement points per answer in this dataset, and proportionally more in the others.
To further demonstrate that \method{} signals can still be predictive of shorter reasoning traces, Figure \ref{300_tokens} demonstrates how our ROC-AUC results are equivalent when considering interval segments of 300 tokens.

In addition, we compared fixed-$k$ strategy to defining segments by newline tokens, as other studies report \citep{sun2025omega}. 
However, segment sizes varied substantially across models under this approach, making it less comparable across architectures.
\clearpage

\section{Models and Inference Settings}
\label{inference_settings}
We perform our inference and evaluation using the \textbf{Eureka ML Insights} framework 
(\href{https://github.com/microsoft/eureka-ml-insights}{\texttt{microsoft/eureka-ml-insights}}).

We used a max generation length of 31,768 tokens for all models.
For all experiments, we report model sources and inference parameters to ensure reproducibility:

\paragraph{DeepSeek-R1-Qwen-14B}  
\href{https://huggingface.co/deepseek-ai/DeepSeek-R1-Distill-Qwen-14B}{\texttt{deepseek-ai/DeepSeek-R1-Distill-Qwen-14B}}  
\begin{itemize}
    \item Temperature = 0.6
    \item Top-$p$ = 0.95  
\end{itemize}

\paragraph{Phi-4-Reasoning-Plus}  
\href{https://huggingface.co/microsoft/Phi-4-reasoning-plus}{\texttt{microsoft/Phi-4-reasoning-plus}}  
\begin{itemize}
    \item Temperature = 0.8
    \item Top-$k$ = 50
    \item Top-$p$ = 0.95
\end{itemize}

\paragraph{Qwen3-14B (thinking enabled)}  
\href{https://huggingface.co/Qwen/Qwen3-14B}{\texttt{Qwen/Qwen3-14B}}  
\begin{itemize}
    \item Temperature = 0.6
    \item Top-$p$ = 0.95
    \item Top-$k$ = 20
\end{itemize}

\end{document}